# Integrated Path Tracking with DYC and MPC using LSTM Based Tire Force Estimator for Four-wheel Independent Steering and Driving Vehicle

Sungjin Lim, Bilal Sadiq, Yongsik Jin, Sangho Lee, Gyeungho Choi, Kanghyun Nam and Yongseob Lim, *Member, IEEE*

***Abstract*— Active collision avoidance system plays a crucial role in ensuring the lateral safety of autonomous vehicles, and it is primarily related to path planning and tracking control algorithms. In particular, the direct yaw-moment control (DYC) system can significantly improve the lateral stability of a vehicle in environments with sudden changes in road conditions. In order to apply the DYC algorithm, it is very important to accurately consider the properties of tire forces with complex nonlinearity for control to ensure the lateral stability of the vehicle. In this study, longitudinal and lateral tire forces for safety path tracking were simultaneously estimated using a long short-term memory (LSTM) neural network based estimator. Furthermore, to improve path tracking performance in case of sudden changes in road conditions, a system has been developed by combining 4-wheel independent steering (4WIS) model predictive control (MPC) and 4-wheel independent drive (4WID) direct yaw-moment control (DYC). The estimation performance of the extended Kalman filter (EKF), which are commonly used for tire force estimation, was compared. In addition, the estimated longitudinal and lateral tire forces of each wheel were applied to the proposed system, and system verification was performed through simulation using a vehicle dynamics simulator. Consequently, the proposed method, the integrated path tracking algorithm with DYC and MPC using the LSTM based estimator, was validated to significantly improve the vehicle stability in suddenly changing road conditions.**

This work was supported by Daegu Gyeongbuk Institute of Science and Technology (DGIST) Institution-Specific Project and supported by the National Research Foundation of Korea (NRF) grant, which is funded by the Korea government (MSIT) via the Ministry of Science and Information and Communications Technology (ICT) of Korea. (No. 22-DPIC-17 and 2021R1F1A1046197), Electronics and Telecommunications Research Institute (ETRI) grant funded by the Korean government (23ZD1160), Development of ICT Convergence Technology for Daegu-Gyeongbuk Regional Industry.

Sungjin Lim, Bilal Sadiq, Gyeungho Choi and Yongseob Lim are with DGIST, 42988 Daegu, South Korea (e-mail: tjd2637@dgist.ac.kr; bilalsadiq@dgist.ac.kr; ghchoi@dgist.ac.kr; yslim73@dgist.ac.kr).

Yongsik Jin is with the Electronics and Telecommunications Research Institute (ETRI), 42995 Daegu, South Korea (e-mail: yongsik@etri.re.kr).

Sangho Lee is with Hyundai Motor Company, 18280 Suwon, South Korea (e-mail: imprince@hyundai.com).

Kanghyun Nam is with Yeungnam university, 38541 Daegu, South Korea (khnam@yu.ac.kr).

Corresponding author: Yongseob Lim.

## I. Introduction

Autonomous vehicles require many reliable functions such as improving passenger comfort, diverse mission accomplishments, road driving safety, and human quality of life [1-5]. Therefore, autonomous driving technology has been considered an important technology that can reduce accidents caused by human errors in the presence of diverse road conditions. As shown in Fig. 1, since vehicle maneuvering stability can be significantly reduced in a low friction road environment owing to rain, snow, ice, and other factors, an autonomous vehicle essentially requires a reliable path tracking controller for stable maneuvering [6].

It is also important to improve the maneuverability of autonomous vehicles to ensure lateral stability, which is an important aspect related to the maneuvering safety of autonomous vehicles. In these studies, the rear wheel steering (RWS), active front steering (AFS), electronic stability control (ESC), and torque vectoring were thoroughly investigated [7-9]. Recently, autonomous vehicles equipped with four-wheel independent driving, steering, and braking systems have been actively studied for improving the maneuvering performance supported by in-wheel motor development [10-12]. Thus, many existing electric vehicles (EVs) tend to use front wheel steering (FWS) or active front steering (AFS) systems. Four-wheel independent steering and driving (4WISD) EVs also tend to use direct yaw-moment control (DYC) technology to provide high-speed maneuverability and handling stability [13].

For the robust and reliable driving performance of autonomous driving, we conducted the proposed research focusing on a LSTM based path tracking control algorithm for autonomous four-wheel independent steering vehicles by using an in-wheel motor application among various autonomous driving technologies, such as localization, mapping, perception, and path planning [4].

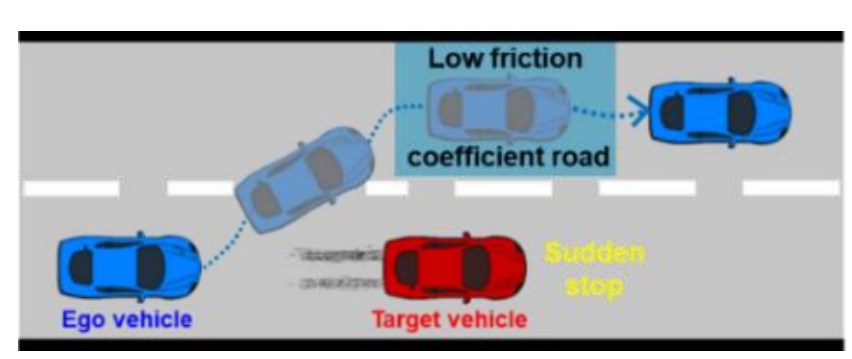

Fig. 1. Overall schematic of the integrated path tracking controller with deep learning learning-based for autonomous vehicle application in the presence of rapid road friction variation.

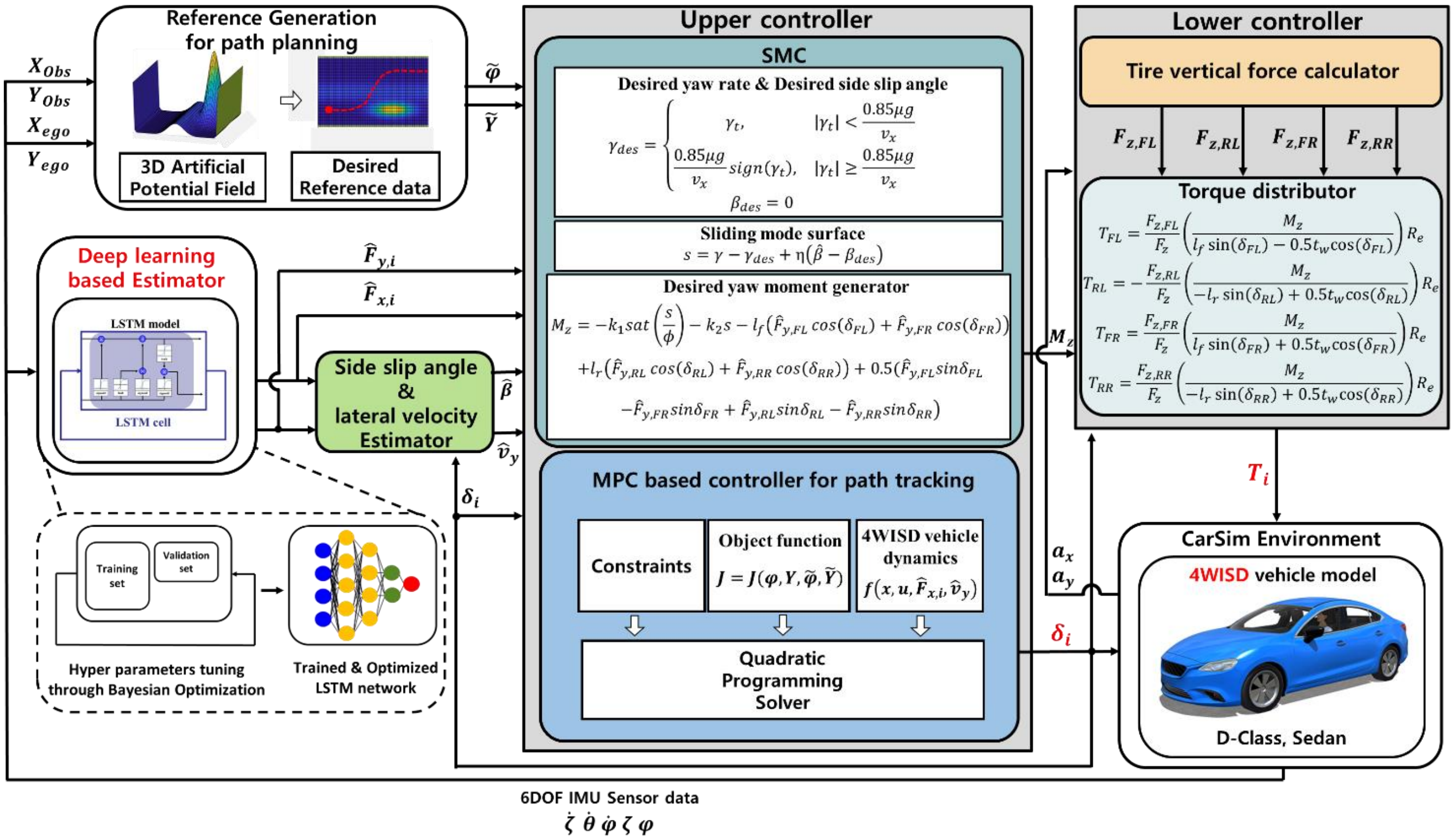

Fig. 2. Overall schematic of the integrated path tracking controller with DYC and MPC using deep learning-based estimator for four-wheel independent steering and driving (4WISD) vehicle.

### *A. Related Work*

Active collision avoidance is a key capability for the safety of autonomous vehicles and is primarily relevant to path planning and tracking control algorithms. Wang et al. constructed a dynamic error model for four-wheel steering (4WS) and analyzed the steady state of a 4WS autonomous vehicle using a sliding mode controller (SMC) [14]. Sun et al. introduced a path tracking control algorithm for autonomous vehicles using a fuzzy model-based $H_\infty$ dynamic output feedback control [15]. Zhang et al. proposed an optimal preview linear quadratic regulator (LQR) based on SMC for path tracking using a bicycle model [16]. The aforementioned controllers exhibit good path tracking performance. However, these studies did not consider vehicle control constraints (e.g., actuator range, rate limits, and safety limits).

The model predictive control (MPC) algorithm, which uses a feedback law computed in real-time by solving a constrained optimal control problem, has been considered as a promising solution for path tracking. Jeong et al. achieved high performance in autonomous driving with high curvature using a linear time-varying vehicle model-based MPC controller [17]. However, the lateral force of the tire was calculated using linear tire model, which inevitably contained model uncertainties. Zheng et al. calculated the tire force required for the MPC algorithm using the Pacejka tire model [18]. This method also includes tire model uncertainties. Thus, to maximize the performance of the 4WIS system, it is important to accurately calculate the longitudinal and lateral tire forces, which are essential for obtaining the optimal steering angles in a linear time-varying MPC-based path tracking controller.

In general, to directly measure the longitudinal and lateral tire forces of a vehicle, a wheel force transducer attached to the inside or outside of each vehicle tire is used [19]. However, direct tire force measurement using sensors is not cost-effective and not practical for vehicle production. Therefore, to overcome the aforementioned disadvantages, a longitudinal and lateral tire force estimator based on a vehicle dynamics model has been extensively investigated. Rezaeian et al. estimated the vertical, longitudinal, and lateral tire forces using an extended Kalman filter (EKF) and an unscented Kalman filter (UKF) [20]. Jung et al. used the EKF with multiple interacting models to estimate the longitudinal and lateral tire forces in all-wheel drive vehicle systems [21]. Moreover, with the recent development of deep learning technology, a data-based estimator for estimating unknown inputs has also been introduced. In a data-driven approach, decisions are made using nonlinear mappings from the input and output data instead of dynamic equations. Therefore, pre-trained models in an offline environment are computationally more efficient than real-time model-based estimators because they only perform the computation of linear activation equations. Im et al. accurately estimated the unknown road input and vertical tire force simultaneously using LSTM. The results showed that the computational efficiency and estimation performance of LSTM significantly improved [22]. Xu et al. estimated the longitudinal and lateral forces on tires using machine learning [23]. However, this was conducted in a quarter-car-based vehicle test unit, not a full vehicle model. This was also estimated by attaching a sensor inside the tire, which is impractical for the vehicle production.

### B. Contribution

The main contribution in this paper are as follows:

- We designed an LSTM estimator that can estimate the longitudinal and lateral tire forces of each wheel using only an inertial measurement unit (IMU) sensor.
- An effectively integrated path tracking controller was designed by applying an LSTM deep learning-based tire force estimator to the MPC and DYC algorithm for a four-wheel independent steering and drive system.
- As the use of multiple sensors in a chassis control system is complex and expensive, the number of sensors should be minimized for securing cost-effectiveness and simplicity. Therefore, we thoroughly validated that LSTM based estimator could be used to reduce the number of measurement sensors compared with the extended Kalman filter, which is commonly used for parameter estimation in vehicle control.

Therefore, the main objective of this study was to design an integrated path tracking controller for significantly improving vehicle stability in cases of emergency avoidance maneuvers using a deep learning-based estimator (i.e., the LSTM network model), which was also combined with the MPC and DYC algorithms, as schematically described in Fig. 2. In addition, we searched for promising hyperparameter sets through Bayesian optimization and used them to design optimal learning models. The local reference trajectory generated by the artificial potential field (APF) considering the location values of the ego vehicle and obstacle was used as a reference for the MPC algorithm [24-25]. A CarSim vehicle dynamics simulator was used to generate the vehicle sensor data used as the training and test data. It was also used to validate the proposed method because it is widely used as an accurate alternative platform for actual vehicle testing [26].

## II. Method

### A. Deep Learning-based Time Series Data Estimator

The longitudinal and lateral tire forces of a vehicle exhibit nonlinear behavior with respect to the slip rate [27]. The designed LSTM model was trained using body data, including the vehicle's nonlinear dynamic behavior as well. Therefore, it is possible to estimate the tire force using the LSTM model by measuring the vehicle body data. The detailed learning process of LSTM is presented in [22, 28]. A flowchart of the forward propagation of LSTM is shown in Fig. 3.

$x$(t) is the input, $h$(t) is the output at the current time step, and l denotes the order of the LSTM cells, respectively. $W$ means the weight at each step. The data are computed using a fully connected layer. Model training was performed to determine all optimal weight values via backpropagation. The hyperparameters required to learn the LSTM model consisted of the maximum epochs, validation frequency, gradient threshold, initial learning rate, learning rate drop period, learning rate drop factor, mini-batch size and sequence length.

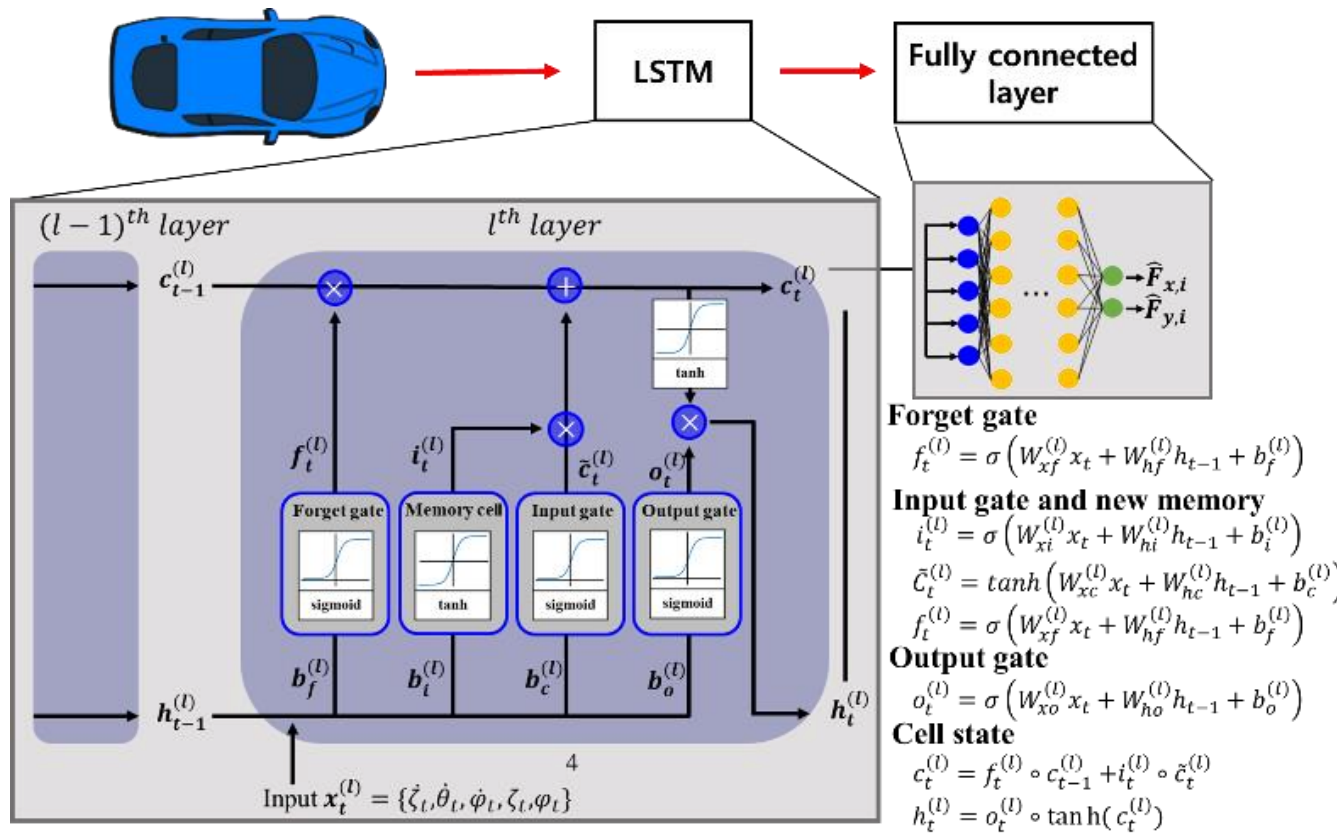

Fig. 3. Schematic block diagram of LSTM forward propagation.

The optimization of the mentioned eight hyperparameters is essential for achieving the best performance in LSTM regression networks. Each hyperparameter plays a vital role in improving the model accuracy, stability, and computational efficiency. By optimizing these hyperparameters, we can obtain a model that efficiently learns from the data, reduces overfitting, and adapts to the training process, while maintaining a balance between resource allocation and gradient estimation. Optimizing maximum epochs ensures that the model is neither underfitted nor overfitted, leading to a well-trained model that generalizes well to new data. Adjusting validation frequency helps monitor the model performance during training, enabling the optimization algorithm to converge faster by identifying overfitting early and guiding its direction. Tuning gradient threshold prevents gradient explosion, enhancing model stability and convergence, resulting in a more accurate and reliable model. Selecting an appropriate initial learning rate impacts on the process speed and stability during training process, ensuring a faster convergence to an optimal solution without oscillations or divergence. Optimizing learning rate drop period and drop factor allows the model to adapt its learning rate during training, improving the convergence rate and avoiding getting stuck in local minima. Choosing the appropriate mini batch size is able to balance resource efficiency and gradient estimation accuracy. The optimization of this parameter ensures that the model trains efficiently without sacrificing the quality of the gradient estimation. Furthermore, optimizing sequence length ensures that the model captures both short-term and long-term dependencies in the data, leading to better model performance while managing computational costs effectively.

### B. Bayesian Optimization for LSTM Based Tire Force Estimator

The Bayesian optimization is a global optimization strategy that stands out from grid search and random search [29-30]. Particularly, grid search exhaustively evaluates user-defined hyperparameter combinations, while random search explores the hyperparameter space randomly. Both approaches can reach optimal results, but they tend to demand significant computational resources. In contrast, Bayesian optimization uses a sample-efficient method, iteratively learning a probabilistic model for optimization problems and selecting promising candidates for the next exploration step. Thus, this approach can result in reduced computational cost and fewer iterations compared to traditional methods in terms of hyperparameter search space, making Bayesian optimization a more efficient optimization technique. Bayesian optimization is operated by simultaneous interaction of the probabilistic surrogate model and the acquisition function to solve the equation represented as equation (1).

$$\lambda^* = \underset{\lambda \in \Omega}{\operatorname{argmin}}\, g(\lambda) \tag{1}$$

The surrogate model probabilistically approximates the unknown objective function $g(\cdot)$, while the acquisition function serves to select the optimal hyperparameter candidates $\lambda_{N+1}$ based on the surrogate model by considering the prior evaluated samples $(\lambda_1, g(\lambda_1)), \ldots, (\lambda_N, g(\lambda_N))$. Following the previously described procedure, the optimal hyperparameter candidate $\lambda_{N+1}$ is selected, and the acquisition function and surrogate model are iteratively updated to identify the optimal $\lambda^*$ value within the entire hyperparameter space. Moreover, Gaussian process is also utilized as a surrogate model, as suggested in [31], and the expected improvement (EI) algorithm introduced in [32] serves as the acquisition function for optimization.

### *C. Training Data Acquisition*

In this study, the LSTM model training process was performed using only the sprung-mass data of the full-vehicle model generated by the CarSim simulation platform. Therefore, a six-dimensional inertial measurement unit (IMU) sensor was used to measure the motion of the sprung mass. The training data was obtained from the 6-axis IMU sensor and consists of roll rate ($\dot{\zeta}$), pitch rate ($\dot{\theta}$), yaw rate ($\gamma$), roll angle ($\zeta$), and yaw angle ($\varphi$). Thus, a learning dataset was obtained by using only the IMU sensor, which is widely used in automation and autonomous vehicle industries, as a cost-effective approach. Next, the LSTM estimator was modeled for training, testing, and verification using 90,000, 20,000, and 10,000 datasets, respectively. Each process was achieved under an environment of driving on a 1.8 km road at a velocity of 80 km/h with a sampling frequency of 1 kHz. Moreover, the friction coefficient of the entire road was 0.85 on asphalt condition, but it was partially composed of a friction coefficient section of 0.2 due to the local pieces of rain, snow, or black ice on the asphalt.

### *D. APF For Obstacle Avoidance Reference Path Generation*

In this study, it is assumed that the ego vehicle travels on a straight road and only a single obstacle exists in front of the moving ego vehicle. In addition, a short potential field within 50 m of the lane change safety distance was implemented to simulate a situation in which the ego vehicle rapidly avoids obstacles in front of it [33]. A potential field is created according to the obstacle and the target position, and guides the vehicle to avoid the obstacle while moving to the target position. At the final target point, the value of the potential field is minimized and directed toward the target point by the attractive potential field. The obstacle potential field has its maximum at the position of the obstacle, causing the vehicle to move away from the obstacle by the repulsive potential field.

The repulsive potential field $P_R(X, Y)$ and attactive potential field $P_A(X)$ are represneted as equations (2) and (3), respectively.

$$P_R(X, Y) = P_o(X, Y) + P_l(Y) + P_r(Y) \tag{2}$$

$$P_A(X) = P_v(X) \tag{3}$$

where $P_o$ is the obstacle potential field, $P_l$ is the lane potential field for lane classification, $P_r$ is road potential field indicating the boundary of the road, $P_v$ is the velocity potential field respectively.

Vehicles are required to maintain a safe distance behind the target vehicle within the lane, whereas it is relatively allowed to be closer to the obstacle on lateral direction. Therefore, the obstacle potential field is represented as equation (4).

$$P_o(X, Y) = A_o e^{-\left\{\frac{(X - X_{obs})^2}{2\sigma_x^2} + \frac{(Y - Y_{obs})^2}{2\sigma_y^2}\right\}} \tag{4}$$

where $A_o$ is the maximum potential field values of obstacle. $X$ and $X_{obs}$ are ego vehicle and obstacle longitudinal positions, respectively. $Y$ and $Y_{obs}$ are ego vehicle and obstacle lateral positions, respectively. $\sigma_x$ and $\sigma_y$ are the standard deviations of the obstacle potential field in the $X$ and $Y$ directions, respectively, and the inclination of the shape can be adjusted.

The center lane potential field is expressed as equation (5), and the left and right road boundary potential field is expressed as equation (6).

$$P_l(Y) = A_l e^{\left\{-\frac{(Y - Y_c)^2}{2\sigma_l{}^2}\right\}} \tag{5}$$

$$P_r(Y) = A_r \left\{ \left( \frac{1}{Y - Y_{lr}} \right)^2 + \left( \frac{1}{Y - Y_{rr}} \right)^2 \right\} \tag{6}$$

where $A_l$ and $A_r$ are the maximum potential field values of lane and road, respectively. $Y_c$ is the lateral position of the lane and $\sigma_l$ is the standard deviations of the lane potential field in the $Y$ direction and the inclination of the shape can be adjusted. $Y_{lr}$ and $Y_{rr}$ are the distance from the center of the lane to the left end of the road and the distance to the right end of the road, respectively.

The purpose of the velocity potential field is to allow the ego vehicle to move forward and is expressed in equation (7).

$$P_v(X) = \gamma_v (v_x - v_d) X \tag{7}$$

where $\gamma_v$ is the slope scale factor, and $v_x$ and $v_d$ are the longitudinal velocity and desired longitudinal velocity, respectively, and $v_d$ is set to be larger than $v_x$. The total potential field $P_T$ is represented by equation (8), and Fig.4 shows the total potential field.

$$P_T(\mathrm{X}, \mathrm{Y}) = P_o + P_l + P_r + P_v \tag{8}$$

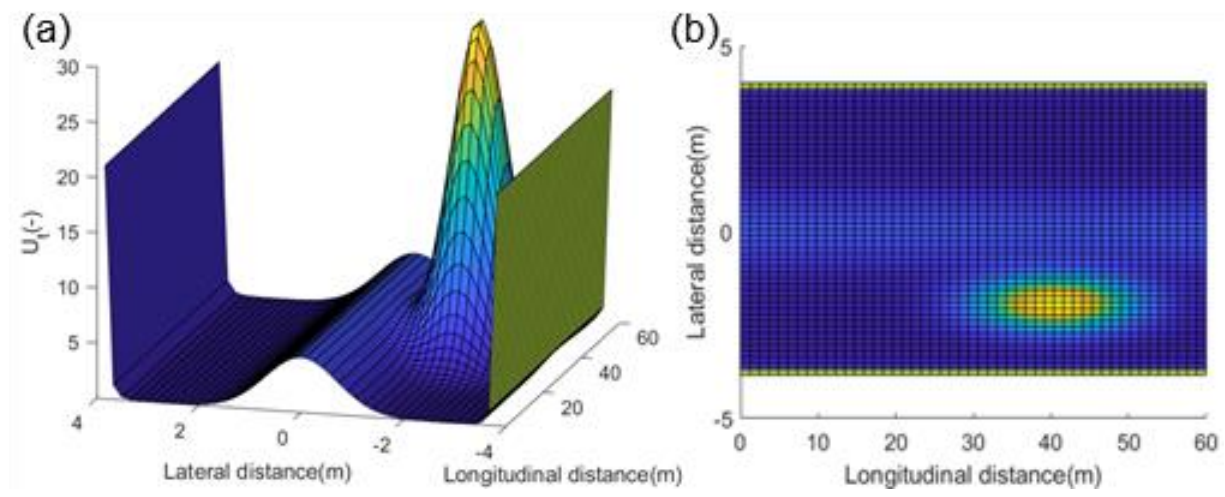

Fig. 4. Total potential field; (a) 3D image, (b) 2D image.

As shown Fig. 5, autonomous vehicle moving by potential fields is attracted by $\vec{F}_A$ and repulsed by $\vec{F}_R$. The total force $\vec{F}_T$ expressed by equation (9) is the sum of the attractive force $\vec{F}_A$ and the repulsive force $\vec{F}_R$, which determines the direction of reasonable movement in the current position.

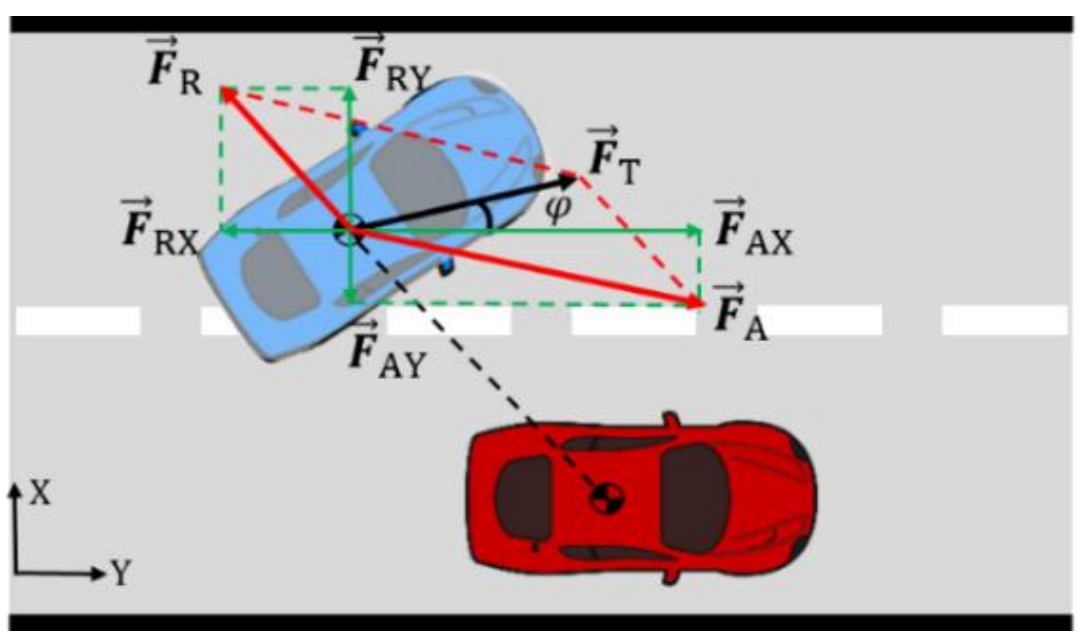

Fig. 5. Direction of the forces by the total potential field.

$$\begin{aligned}\vec{F}_T &= -\nabla P_T(X,Y) = -\nabla P_A(X,Y) - \nabla P_R(X,Y) \\ &= \left(\vec{F}_{AX} + \vec{F}_{RX}\right) + \left(\vec{F}_{AY} + \vec{F}_{RY}\right) \\ &= -\frac{\partial P_T(X,Y)}{\partial X} - \frac{\partial P_T(X,Y)}{\partial Y}\end{aligned} \tag{9}$$

where $\vec{F}_{AX}$ and $\vec{F}_{RX}$ are the attractive force and repulsive force in $X$ direction. $\vec{F}_{AY}$ and $\vec{F}_{RY}$ are the attractive force and repulsive force in $Y$ direction. Therefore, the $X$ and $Y$ components of $\vec{F}_T$ can be expressed as follows:

$$\begin{aligned}\vec{F}_T &= -\nabla P_T(X,Y) = -\begin{bmatrix}\dfrac{\partial P_T(X,Y)}{\partial X} \\ \dfrac{\partial P_T(X,Y)}{\partial Y}\end{bmatrix} = \begin{bmatrix}\vec{F}_{TX} \\ \vec{F}_{TY}\end{bmatrix} \\ &= -\begin{bmatrix}\dfrac{\partial\left(P_o(X,Y) + P_v(X)\right)}{\partial X} \\ \dfrac{\partial\left(P_o(X,Y) + P_l(Y) + P_r(Y)\right)}{\partial Y}\end{bmatrix}\end{aligned} \tag{10}$$

The yaw angle φ which is required to follow the local path generated by the potential field from the current position can be calculated as shown in equation (11).

$$\begin{cases}\varphi = tan^{-1}\left(\dfrac{\vec{F}_{TY}}{\vec{F}_{TX}}\right), \vec{F}_{TX} > 0 \\ \varphi = \pi + tan^{-1}\left(\dfrac{\vec{F}_{TY}}{\vec{F}_{TX}}\right), \vec{F}_{TX} \le 0\end{cases} \tag{11}$$

Using φ calculated from equation (11), the vehicle global position can be expressed as follows:

$$\begin{aligned}X &= X_0 + \int v_x cos(\varphi) - v_y cos(\varphi) dt \\ Y &= Y_0 + \int v_x sin(\varphi) + v_y cos(\varphi_t) dt\end{aligned} \tag{12}$$

where $X_0$ and $Y_0$ are the initial positions in the $X$ and $Y$ directions, respectively. The yaw angle φand its corresponding dependent global $Y$ coordinate are used as references in the MPC controller.

*E. Four-wheel Independent Steering Vehicle Model*

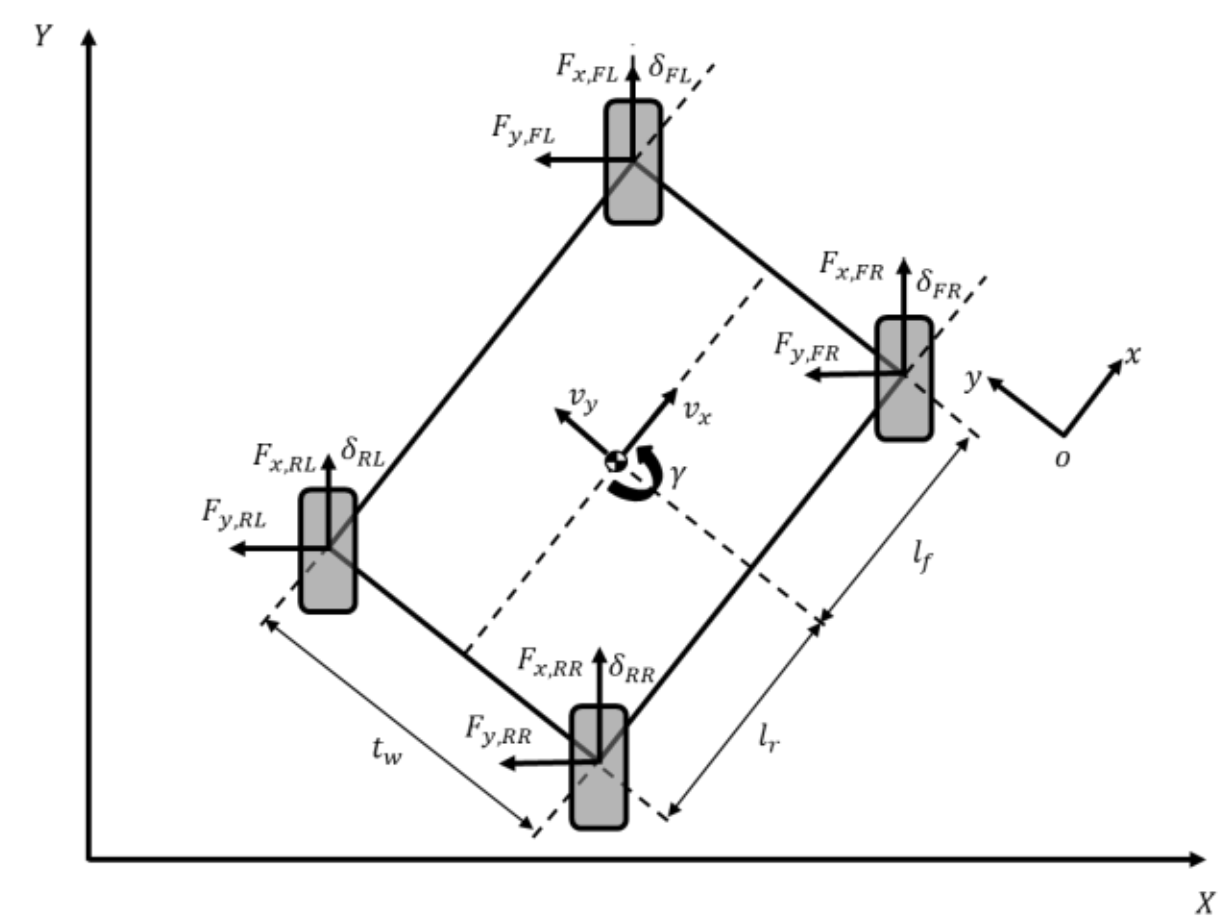

Fig. 6. Schematic of 4WIS vehicle system model.

To model the 4WIS vehicle system, three dynamic models of the vehicle longitudinal, lateral, and yaw directions were used, as shown in Fig. 6. The behavior of the full vehicle model is defined in three directions, where *x* is the longitudinal direction of the vehicle traveling direction, *y* is the lateral direction of the vehicle turning direction, and *z* is the vertical direction perpendicular to *x* and *y*. Moreover, the rotation of the full vehicle model around each of the *x*, *y*, and *z* direction axes is defined as the roll, pitch, and yaw motion, respectively, and is represented by equations (13), (14), and (15).

$$\begin{aligned}m\dot{v}_x = v_y\gamma + F_{x,FL}cos\delta_{FL} - F_{y,FL}sin\delta_{FL} + \\ F_{x,FR}cos\delta_{FR} - F_{y,FR}sin\delta_{FR} + F_{x,RL}cos\delta_{RL} - \\ F_{y,RL}sin\delta_{RL} + F_{x,RR}cos\delta_{RR} - F_{y,RR}sin\delta_{RR}\end{aligned} \tag{13}$$

$$\begin{aligned}m\dot{v}_y = -v_x\gamma + F_{x,FL}sin\delta_{FL} + F_{y,FL}cos\delta_{FL} + \\ F_{x,FR}sin\delta_{FR} + F_{y,FR}cos\delta_{FR} + F_{x,RL}sin\delta_{RL} + \\ F_{y,RL}cos\delta_{RL} + F_{x,RR}sin\delta_{RR} + F_{y,RR}cos\delta_{RR}\end{aligned} \tag{14}$$

$$\begin{aligned}I_z\dot{\gamma} = l_f\left(F_{y,FL}cos\delta_{FL} + F_{y,FR}cos\delta_{FR}\right) - \\ l_r\left(F_{y,RL}cos\delta_{RL} + F_{y,RR}cos\delta_{RR}\right) - 0.5t_w\left(F_{y,FL}sin\delta_{FL} - \right. \\ \left. F_{y,FR}sin\delta_{FR} + F_{y,RL}sin\delta_{RL} - F_{y,RR}sin\delta_{RR}\right) + M_z\end{aligned} \tag{15}$$

where $m$ is mass of the vehicle body and $I_z$ is the yaw moment of inertia. $v_x$, $v_y$ and γ are the longitudinal, lateral velocity and yaw rate of vehicle respectively; and $F_{x,i}(i = FL, FR, RL, RR)$ and $F_{y,i}(i = FL, FR, RL, RR)$ are longitudinal and lateral tire forces of each wheel, respectively; and $\delta_i(i = FL, FR, RL, RR)$ are steering angle of each wheel, respectively; $l_f$, are forward length and backward length of vehicle respectively and $t_w$ is width of vehicle. $M_z$ is used as the yaw-moment controller and is expressed by equation (16).

$$M_z = l_f\left(F_{x,FL}sin\delta_{FL} + F_{x,FR}sin\delta_{FR}\right) - l_r\left(F_{x,RL}sin\delta_{RL} + F_{x,RR}sin\delta_{RR}\right) - 0.5t_w\left(-F_{x,FL}cos\delta_{FL} + F_{x,FR}cos\delta_{FR} - F_{x,RL}cos\delta_{RL} + F_{x,RR}cos\delta_{RR}\right) \quad (16)$$

The physical parameters of the 4WISD model are listed in Table I.

TABLE I
PHYSICAL PARAMETER THE 4WISD SYSTEM MODEL
(D-CLASS, SEDAN)

| Symbol | Parameter | Value | Unit |
|---|---|---|---|
| $m$ | Vehicle mass | 1685.2 | $kg$ |
| $I_z$ | Yaw moment of inertia | 2315.3 | $kg{\cdot}m^2$ |
| $I_w$ | Wheel moment of inertia | 1.5 | $kg{\cdot}m^2$ |
| $t_w$ | Vehicle width | 1.795 | $m$ |
| $l_f$ | Vehicle length (forward) | 1.110 | $m$ |
| $l_r$ | Vehicle length (backward) | 1.756 | $m$ |
| $C_f$ | Front tire cornering stiffness | 46235 | $N/rad$ |
| $C_r$ | Rear tire cornering stiffness | 31442 | $N/rad$ |
| $R_r$ | Unloaded wheel radius | 0.325 | $m$ |
| $R_e$ | Effective wheel rolling radius | 0.334 | $m$ |
| $\Delta t$ | Sampling time | 0.01 | *second* |

Assuming that the tire slip angle is small, the tire lateral force is linearly expressed as the product of cornering stiffness and cornering angle, as shown in equations (17-20). Here, $C_f$ and $C_r$ are the cornering stiffness of the front and rear wheels; and $\alpha_i(i = FL, FR, RL, RR)$ is the slip angle of each tire.

$$F_{y,FL} = C_f\hat{\alpha}_{FL} = C_f\left(\delta_{FL} - \frac{\hat{v}_y + l_f\gamma}{v_x}\right) \quad (17)$$

$$F_{y,FR} = C_f\hat{\alpha}_{FR} = C_f\left(\delta_{FR} - \frac{\hat{v}_y + l_f\gamma}{v_x}\right) \quad (18)$$

$$F_{y,RL} = C_r\hat{\alpha}_{RL} = C_r\left(\delta_{RL} - \frac{\hat{v}_y - l_r\gamma}{v_x}\right) \quad (19)$$

$$F_{y,RR} = C_r\hat{\alpha}_{RR} = C_r\left(\delta_{RR} - \frac{\hat{v}_y - l_r\gamma}{v_x}\right) \quad (20)$$

In order to calculate the accurate lateral tire force which are required for the MPC model, it is crucial to precisely compute the lateral vehicle velocity $\hat{v}_y$. By substituting the estimated longitudinal and lateral tire force values obtained from the LSTM into the side slip dynamics from equation (21), it becomes possible to calculate the side slip angle. Then, lateral vehicle velocity is obtained from equation (22).

$$mv_x\dot{\hat{\beta}} = -\hat{F}_{x,FL}sin(\hat{\beta} - \delta_{FL}) - \hat{F}_{x,FR}\sin(\hat{\beta} - \delta_{FR}) - \hat{F}_{x,RL}\sin(\hat{\beta} - \delta_{RL}) - \hat{F}_{x,RR}\sin(\hat{\beta} - \delta_{RR}) + \hat{F}_{y,FL}cos(\delta_{FL} - \hat{\beta}) + \hat{F}_{y,FR}cos(\delta_{FR} - \hat{\beta}) + \hat{F}_{y,RL}cos(\delta_{RL} - \hat{\beta}) + \hat{F}_{y,RR}cos(\delta_{RR} - \hat{\beta}) \quad (21)$$

$$\hat{v}_y = v_x\tan\left(\hat{\beta}\right) \quad (22)$$

The non-linear state space equation of 4WIS model and the output equation can be expressed as equation (23) and the state function $f$ of each model and the measurement function $h$ can be expressed as equation (24) and (25), respectively.

$$\dot{x}(t) = f\left(x(t), u(t)\right) \quad (23)$$

$$y^*(t) = h\left(x(t), u(t)\right)$$

The state vector $x$, output vector $y$, and input vector $u$ used for the MPC model are defined in equation (26). To apply to linear time varying MPC controller, equation (27) expressed as a continuous time equation should be linearized considering the current state $x$(t) and input $u$(t-1).

$$\begin{gathered} x = [v_x \quad v_y \quad \gamma \quad \varphi \quad Y] \\ y = [\varphi \quad Y] \\ u = [\delta_i] \end{gathered} \quad (26)$$

where $\delta_i(i = FL, FR, RL, RR)$ is road wheel steering angle of each wheel. The state-space equation derived in equation (23) for the linear time varying MPC controller can be expressed as follows:

$$\begin{gathered} \dot{x}(t) = A(t)x(t) + B(t)u(t) \\ y(t) = C(t)x(t) \end{gathered} \quad (27)$$

where

$$A(t) = \frac{\partial f\left(x(t), u(t)\right)}{\partial x}, B(t) = \frac{\partial f\left(x(t), u(t)\right)}{\partial u}$$

$$C(t) = \frac{\partial h\left(x(t), u(t)\right)}{\partial x}$$

The first-order difference method is applied to convert the linear time-varying continuous system in (27) into its discrete-time equivalent, as represented by equation (28).

$$\begin{gathered} x(k) = A_dx(k-1) + B_du(k-1) \\ y(k-1) = Cx(k-1) + Du(k-1) \end{gathered} \quad (28)$$

where $A_d = I_5 + A(t)\Delta\text{t}$ is the discretized state matrix and $B_d = B(t)\Delta\text{t}$ is discretized input matrix.

## III. CONTROLLER DESIGN

### A. Linear Time-varying Model Predictive Controller

The MPC algorithm used in this study uses a discretized state-space equation. The control input can be calculated by the definition in equation (29). The augmented state-space equation used in MPC consists of an augmented state vector $\tilde{x}(k)$ and an increment input $\Delta u(k)$ defined in equation (30).

$$u(k+1) = u(k) + \Delta u(k) \quad (29)$$

$$\begin{cases} \tilde{x}(k+1) = A\tilde{x}(k) + B\Delta u(k) \\ \tilde{y}(k) = C\tilde{x}(k) \end{cases} \quad (30)$$

where $A = \begin{bmatrix} A_d & B_d \\ 0_{4\times5} & I_4 \end{bmatrix}$ is the state matrix, $B = [B_d \quad I_4]^T$ is the input matrix, $\text{C} = [C_d \quad 0_{2\times4}]$ is the output matrix, $\tilde{x} = [x(k) \quad u(k)]^T$ is the augmented state vector, $A_d = I_4 + \Delta tA_c(t)$ is the discretized state matrix and $B_d = \Delta tB_c(t)$ is the discretized input matrix, respectively. Then, the MPC algorithm was used to calculate a prediction output that was as close as possible to the reference trajectory within the prediction horizon, whereas the prediction output was used to compute an optimal control input. For reference tracking purposes, the cost function of the MPC problem is defined in the quadratic form in equation (31) and in vector form as defined in equation (32).

$$
\begin{cases}
f_1 = \frac{1}{m}\{\hat{F}_{x,FL}\cos(u_1) - C_{FL}\sin(u_1)\left(u_1 - \left(\frac{\hat{v}_y + l_f x_3}{v_x}\right)\right) + \hat{F}_{x,FR}\cos(u_2) - C_{FR}\sin(u_2)\left(u_2 - \left(\frac{\hat{v}_y + l_f x_3}{v_x}\right)\right) \\
+\hat{F}_{x,RL}\cos(u_3) - C_{RL}\sin(u_3)\left(u_3 - \left(\frac{\hat{v}_y - l_r x_3}{v_x}\right)\right) + \hat{F}_{x,RR}\cos(u_4) - C_{RR}\sin(u_4)\left(u_4 - \left(\frac{\hat{v}_y - l_r x_3}{v_x}\right)\right)\} + x_2 x_3 \\
f_2 = \frac{1}{m}\{\hat{F}_{x,FL}\sin(u_1) + C_{FL}\cos(u_1)\left(u_1 - \left(\frac{\hat{v}_y + l_f x_3}{v_x}\right)\right) + \hat{F}_{x,FR}\sin(u_2) + C_{FR}\cos(u_2)\left(u_2 - \left(\frac{\hat{v}_y + l_f x_3}{v_x}\right)\right) \\
+\hat{F}_{x,RL}\sin(u_3) + C_{RL}\cos(u_3)\left(u_3 - \left(\frac{\hat{v}_y - l_r x_3}{v_x}\right)\right) + \hat{F}_{x,RR}\sin(u_4) + C_{RR}\cos(u_4)\left(u_4 - \left(\frac{\hat{v}_y - l_r x_3}{v_x}\right)\right)\} - x_1 x_3 \\
f_3 = \frac{1}{I_z}(f_{3^*} + f_{2^*}) \\
f_{3^*} = l_f\left(\hat{F}_{x,FL}\sin(u_1) + C_{FL}\cos(u_1)\left(u_1 - \left(\frac{\hat{v}_y + l_f x_3}{v_x}\right)\right) + \hat{F}_{x,FR}\sin(u_2) + C_{FR}\cos(u_2)\left(u_2 - \left(\frac{\hat{v}_y + l_f x_3}{v_x}\right)\right)\right) \\
-l_r\left(\hat{F}_{x,RL}\sin(u_3) + C_{RL}\cos(u_3)\left(u_3 - \left(\frac{\hat{v}_y - l_r x_3}{v_x}\right)\right) + \hat{F}_{x,RR}\sin(u_4) + C_{RR}\cos(u_4)\left(u_4 - \left(\frac{\hat{v}_y - l_r x_3}{v_x}\right)\right)\right) \\
f_{2^*} = 0.5t_w\left(\hat{F}_{x,FL}\cos(u_1) - \hat{F}_{x,FR}\cos(u_2) + \hat{F}_{x,RL}\cos(u_3) - \hat{F}_{x,RR}\cos(u_4)\right) - C_{FL}\sin(u_1)\left(u_1 - \left(\frac{\hat{v}_y + l_f x_3}{v_x}\right)\right) \\
+C_{FR}\sin(u_2)\left(u_2 - \left(\frac{\hat{v}_y + l_f x_3}{v_x}\right)\right) - C_{RL}\sin(u_3)\left(u_3 - \left(\frac{\hat{v}_y - l_r x_3}{v_x}\right)\right) + C_{RR}\sin(u_4)\left(u_4 - \left(\frac{\hat{v}_y - l_r x_3}{v_x}\right)\right) \\
f_4 = x_3 \\
f_5 = x_2
\end{cases} \tag{24}
$$

$$
\begin{cases} h_1 = x_4 \\ h_2 = x_5 \end{cases} \tag{25}
$$

---

$$
J = \sum_{j=N_1}^{N_p}\left[r(k+j) - y_p(k+j)\right]^T Q\left[r(k+j) - y_p(k+j)\right] + \sum_{j=0}^{N_u-1}\Delta u^T(k) R \Delta u(k) \tag{31}
$$

$$
J = [\Xi(k) - Y_p(k)]^T \bar{Q}\left[\Xi(k) - Y_p(k)\right] + \Delta U^T(k)\bar{R}\Delta U(k) \tag{32}
$$

where $y_p$ is the predicted value, $r$ is the reference value, $\Xi$ is yaw angle φ and global position $Y$ reference sequences generated by the potential field, $Y_p$is the predicted value of output sequence.

$$
\Xi(k) = \begin{bmatrix} r(k+N_1) \\ \vdots \\ r(k+N_p) \end{bmatrix}, Y_p(k) = \begin{bmatrix} y_p(k+N_1) \\ \vdots \\ y_p(k+N_p) \end{bmatrix}
$$

$$
\Delta U(k) = \begin{bmatrix} \Delta u(k) \\ \vdots \\ \Delta u(k+N_u-1) \end{bmatrix}
$$

$$
\bar{Q} = diag\{Q, \cdots, Q\}, \bar{Q} = diag\{R, \cdots, R\}
$$

The predicted output $Y_p$ derived from equation (31) is expressed in equation (33).

$$
Y_p(k) = F\tilde{x}(k) + H\Delta U(k) \tag{33}
$$

where $F = \begin{bmatrix} CA^{N_1} \\ \vdots \\ CA^{N_p} \end{bmatrix}, H = \begin{bmatrix} h_{N_1,1} & \cdots & h_{N_1,N_u} \\ \vdots & \ddots & \vdots \\ h_{N_p,1} & \cdots & h_{N_p,N_u} \end{bmatrix}$ and $h_{j,i} = \begin{cases} CA^{j-i}B, j \geq i \\ 0, j < i \end{cases}$.

Considering the predicted $Y$-value in equations (31) and (33), the cost function can be expressed as in equation (34).

$$
J = \Delta U^T(k)[H^T\bar{Q}H + \bar{R}]\Delta U(k) - 2[\Xi(k) - f(k)]^T\bar{Q}H\Delta U(k) \tag{34}
$$

The constraints were the mean values of adjusting the output values such that the vehicle maneuvered in a stable state and operated the actuator only within the operable range. Three constraints were considered in this study. The input constraints, which can be determined in the range of the size of the input, rate of change of the input, and constraints of the calculated output, are shown in equations (35), (36), and (37), respectively.

$$
-u_s(k) \leq u(k) \leq u_s(k) \tag{35}
$$

$$
-\Delta u_s(k) \leq \Delta u(k) \leq \Delta u_s(k) \tag{36}
$$

$$
-Y(k) \leq Y(k) \leq Y(k) \tag{37}
$$

The prediction horizon $N_p$, control horizon $N_u$, output error weight $Q$, and control increment weight $R$ were 16, 4, 1, and 1000, respectively. The magnitude of the input $u_s$, the change rate of the input $\Delta u_s$ and the magnitude of the output (i.e., yaw $\varphi$ and global $Y$ position $Y_g$) were 21 deg, 90 deg/s, 21 deg, and 5 m, respectively. Finally, optimal $\Delta U^*$, as expressed in equation (38), to minimize the cost function, as expressed in equation (34) by considering the constraints, could be calculated in a quadratic programming (QP) problem.

$$\Delta U^* = \underset{\Delta U(k)}{\operatorname{argmin}}\{\Delta U^T(k)[H^T\bar{Q}H + \bar{R}]\Delta U(k) - 2[\Xi(k) - f(k)]^T\bar{Q}H\Delta U(k)\} \quad (38)$$

*B. Upper-level Controller*

SMC is a well-known nonlinear control method for dealing with uncertain nonlinear systems, parametric uncertainties, and external disturbances [34]. Following the standard design approach of a traditional SMC, the sliding mode surface is defined by equation (39).

$$s = \gamma - \gamma_{des} + \eta(\hat{\beta} - \beta_{des}) \quad (39)$$

where the weight coefficient $\eta$>0 indicates the relative impact of the side slip angle deviation, and its value has been set to 0.01. $\gamma_t$ and $\beta_{des}$ denote desired yaw rate and desired side slip angle for 4WIS vehicle model, as expressed in equations (40) and (41), respectively [34].

$$\gamma_t = \frac{1}{1+\tau_\gamma}\gamma_o \quad (40)$$

$$\beta_{des} = 0 \quad (41)$$

where $\tau_\gamma$ is the time constant, $\gamma_o$ denote desired yaw rate for FWS bicycle model, as expressed in equation (42) [36].

$$\gamma_o = \frac{v_x}{l_f + l_r + \frac{mv_x^2(l_rC_r - l_fC_f)}{2C_rC_f(l_f+l_r)}}\delta_F \quad (42)$$

where $\delta_F = (\delta_{FL} + \delta_{FR})/2$ is the average front steering wheel angle. The desired yaw rate should be bounded by considering the friction coefficient between the tire and the road, as expressed in equation (43).

$$\gamma_{des} = \begin{cases} \gamma_t, & |\gamma_t| < \frac{0.85\mu g}{v_x} \\ \frac{0.85\mu g}{v_x}sign(\gamma_t), & |\gamma_t| \geq \frac{0.85\mu g}{v_x} \end{cases} \quad (43)$$

By differentiating equation (39) and combining equation (15), it can be represented as follows:

$$\dot{s} = \frac{1}{I_z}\Big[l_f(\hat{F}_{y,FL}cos\delta_{FL} + \hat{F}_{y,FR}cos\delta_{FR}) - l_r(\hat{F}_{y,RL}cos\delta_{RL} + \hat{F}_{y,RR}cos\delta_{RR}) - 0.5t_w(\hat{F}_{y,FL}sin\delta_{FL} - \hat{F}_{y,FR}sin\delta_{FR} + \hat{F}_{y,RL}sin\delta_{RL} - \hat{F}_{y,RR}sin\delta_{RR}) + M_z + I_z(\eta(\dot{\hat{\beta}} - \dot{\beta}_{des}) - \dot{\gamma}_{des})\Big] \quad (44)$$

Here, the $\hat{\beta}$ changes slowly and the $\gamma_{des}$ is bound as previously defined, so that the $\dot{\gamma}_{des}$ is also bounded. Therefore, it can be seen that $P$ is a constant and satisfies the condition $\left|I_z(\eta(\dot{\hat{\beta}} - \dot{\beta}_{des}) - \dot{\gamma}_{des})\right| \leq P$.

In a traditional SMC, chattering occurs when the sliding plane $s$ intersects with the control discontinuity points. To decrease this chattering, the sliding mode yaw moment controller $M_z$ is designed with consideration of the boundary layer thickness ($\phi$) as described in equation (45).

$$\begin{aligned} M_z = &-k_1 sat\left(\frac{s}{\phi}\right) - k_2 s \\ &-l_f(\hat{F}_{y,FL}\,cos(\delta_{FL}) + \hat{F}_{y,FR}\,cos(\delta_{FR})) \\ &+l_r(\hat{F}_{y,RL}\,cos(\delta_{RL}) + \hat{F}_{y,RR}\,cos(\delta_{RR})) \\ &+0.5(\hat{F}_{y,FL}sin\delta_{FL} - \hat{F}_{y,FR}sin\delta_{FR} \\ &\hat{F}_{y,RL}sin\delta_{RL} - \hat{F}_{y,RR}sin\delta_{RR}) \end{aligned} \quad (45)$$

with $k_1 > P$ and $k_2 > 0$, then along the sliding surface with $s = 0$, the yaw rate and side slip angle error terms of $s$ converge to the origin and the proof can be derived as follows:

By substituting equation (45) into equation (44), it is expressed as equation (46).

$$\dot{s} = \frac{1}{I_z}\left[-k_1 sat\left(\frac{s}{\phi}\right) - k_2 s + I_z(\eta(\dot{\hat{\beta}} - \dot{\beta}_{des}) - \dot{\gamma}_{des})\right] \quad (46)$$

A Lyapunov function is defined as $V = \frac{1}{2}s^2$, and its time derivative can be expressed as equation (47).

$$\begin{aligned} \dot{V} &= \frac{-k_1 sat\left(\frac{s}{\phi}\right)s - k_2 s^2 + s(\eta(\dot{\hat{\beta}} - \dot{\beta}_{des}) - \dot{\gamma}_{des})}{I_z} \\ &\leq \frac{-k_1\left|sat\left(\frac{s}{\phi}\right)\right|s - k_2 s^2 + |s|\left|(\eta(\dot{\hat{\beta}} - \dot{\beta}_{des}) - \dot{\gamma}_{des})\right|}{I_z} \\ &< 0 \end{aligned} \quad (47)$$

with $k_1 > 0, k_2 > 0$.

The stability of the DYC system, designed based on the finite-time Lyapunov stability theory presented in [33], has been proven.

*C. Lower-level Controller*

The torque distribution equation for the 4WIDS system is derived from equation (48), representing the relationship between motor torque and longitudinal tire force.

$$F_{x,i} = \frac{T_i}{R_e} \quad (48)$$

where $T_i(i = FL, FR, RL, RR)$ is the torque of each wheel.

Taking into account the steering and torque of each wheel, the torque distribution equation is described as equation (49) in order to generate the yaw moment calculated by the upper-level controller.

$$\begin{aligned} T_{FL} &= \frac{F_{z,FL}}{F_z}\left(\frac{M_z}{l_f\sin(\delta_{FL}) - 0.5t_w\cos(\delta_{FL})}\right)R_e \\ T_{RL} &= -\frac{F_{z,RL}}{F_z}\left(\frac{M_z}{-l_r\sin(\delta_{RL}) + 0.5t_w\cos(\delta_{RL})}\right)R_e \\ T_{FR} &= \frac{F_{z,FR}}{F_z}\left(\frac{M_z}{l_f\sin(\delta_{FR}) + 0.5t_w\cos(\delta_{FR})}\right)R_e \\ T_{RR} &= \frac{F_{z,RR}}{F_z}\left(\frac{M_z}{-l_r\sin(\delta_{RR}) + 0.5t_w\cos(\delta_{RR})}\right)R_e \end{aligned} \quad (49)$$

where $F_z$ is the sum of the vertical forces of each wheel and the torque distributed to each wheel is multiplied by the normalized vertical load $F_{z,i}/F_z (i = FL, FR, RL, RR)$, which serves as an indicator for torque distribution. $F_{z,\mathrm{i}}$ is calculated using the equation presented in [37].

## IV. Performance validation

### *A. Extended Kalman Filter Tire Force Estimator Design*

In this study, an EKF was also developed for tire force estimation to verify the performance of the LSTM based estimator. The vehicle model is the same as the previously introduced 4WISD vehicle model. The lateral tire force was calculated using the first-order dynamic tire model presented in [21]. The nonlinear state-space equation and output equation for each model can be expressed as follows:

$$\begin{aligned} \dot{x}^*(t) &= f\big(x^*(t), u^*(t)\big) \\ y^*(t) &= h\big(x^*(t), u^*(t)\big) \end{aligned} \tag{50}$$

where the state vector $x^*$, output vector $y^*$, and input vector $u^*$ used for the extended Kalman filter are defined in equation (51).

$$\begin{aligned} x^* &= [v_x \quad v_y \quad \gamma \quad w_i \quad F_{x,i} \quad F_{y,i}]^T \\ &= [x_1\ x_2\ \cdots\ x_{15}]^T \\ y^* &= [v_x \quad v_y \quad a_x \quad a_y \quad \gamma \quad w_i]^T \\ u^* &= [\delta_i \quad T_i]^T \end{aligned} \tag{51}$$

where $T_i (i = FL, FR, RL, RR)$ is a wheel torque of each wheel. The state function $f^*$ of each model and the measurement function $h^*$ can be expressed as follows:

$$\begin{cases} f_1^* = \dfrac{1}{m}\{x_8\cos(u_1) - x_{12}\sin(u_1) + x_9\cos(u_2) - x_{13}\sin(u_2) \\ \quad +x_{10}\cos(u_3) - x_{14}\sin(u_3) + x_{11}\cos(u_4) - x_{15}\sin(u_4)\} + x_2x_3 \\ f_2^* = \dfrac{1}{m}\{x_8\sin(u_1) + x_{12}\cos(u_1) + x_9\sin(u_2) + x_{13}\cos(u_2) \\ \quad +x_{10}\sin(u_3) + x_{14}\cos(u_3) + x_{11}\sin(u_4) + x_{15}\cos(u_4)\} - x_1x_3 \\ f_3^* = \dfrac{1}{I_z}\{l_f(x_8\sin(u_1) + x_{12}\cos(u_1) + x_9\sin(u_2) + x_{13}\cos(u_2)) \\ \quad -l_r(x_{10}\sin(u_3) + x_{14}\cos(u_3) + x_{11}\sin(u_4) + x_{15}\cos(u_4)) \\ \quad -0.5t_w(-x_8\cos(u_1) + x_9\cos(u_2) - x_{10}\cos(u_3) + x_{11}\cos(u_4) \\ \quad +x_{12}\sin(u_1) - x_{13}\sin(u_2) + x_{14}\sin(u_3) - x_{15}\sin(u_4))\} \\ f_4^* = \dfrac{1}{I_w}\Big(u_5 - R_e\big(x_8 + R_rF_{z,FL}\big)\Big) \\ f_5^* = \dfrac{1}{I_w}\Big(u_6 - R_e\big(x_9 + R_rF_{z,FR}\big)\Big) \\ f_6^* = \dfrac{1}{I_w}\Big(u_7 - R_e\big(x_{10} + R_rF_{z,RL}\big)\Big) \\ f_7^* = \dfrac{1}{I_w}\Big(u_8 - R_e\big(x_{11} + R_rF_{z,RR}\big)\Big) \\ f_8^* = 0, \quad f_9^* = 0, \quad f_{10}^* = 0, \quad f_{11}^* = 0, \\ f_{12}^* = \dfrac{x_1}{\sigma}\big(-x_{12} + \bar{F}_{y,FL}\big), \quad f_{13}^* = \dfrac{x_1}{\sigma}\big(-x_{13} + \bar{F}_{y,FR}\big) \\ f_{14}^* = \dfrac{x_1}{\sigma}\big(-x_{14} + \bar{F}_{y,RL}\big), \quad f_{15}^* = \dfrac{x_1}{\sigma}\big(-x_{15} + \bar{F}_{y,RR}\big) \end{cases}$$

$$\begin{cases} h_1^* = x_1, \quad h_2^* = x_2 \\ h_3^* = \dfrac{1}{m}\{x_8\cos(u_1) - x_{12}\sin(u_1) + x_9\cos(u_2) - x_{13}\sin(u_2) \\ \quad x_{10}\cos(u_3) - x_{14}\sin(u_3) + x_{11}\cos(u_4) - x_{15}\sin(u_4)\} \\ h_4^* = \dfrac{1}{m}\{x_8\sin(u_1) + x_{12}\cos(u_1) + x_9\sin(u_2) + x_{13}\cos(u_2) \\ \quad +x_{10}\sin(u_3) + x_{14}\cos(u_3) + x_{11}\sin(u_4) + x_{15}\cos(u_4)\} \\ h_5^* = x_3, \quad h_6^* = x_4, \quad h_7^* = x_5, \quad h_8^* = x_6, \quad h_9^* = x_7 \end{cases}$$

The state-space equation of the EKF derived in equation (50) for the 4WIS vehicle model can be expressed as follows:

$$\begin{aligned} \dot{x}(t) &= A^*(t)x^*(t) + B^*(t)u^*(t) \\ y(t) &= C^*(t)x^*(t) + Du^*(t) \end{aligned} \tag{51}$$

where,

$$\begin{aligned} A^*(t) &= \frac{\partial f\big(x^*(t), u^*(t)\big)}{\partial x^*}, B^*(t) = \frac{\partial f\big(x^*(t), u^*(t)\big)}{\partial u^*} \\ C^*(t) &= \frac{\partial h\big(x^*(t), u^*(t)\big)}{\partial x^*}, D^*(t) = \frac{\partial h\big(x^*(t), u^*(t)\big)}{\partial u^*} \end{aligned}$$

The first-order difference method is applied to convert the linear time-varying continuous system in (51) into its discrete-time equivalent, as represented by equation (52).

$$\begin{aligned} x^*(k) &= A_d^*x^*(k-1) + B_d^*u^*(k-1) \\ y^*(k-1) &= C^*x^*(k-1) + D^*u^*(k-1) \end{aligned} \tag{52}$$

where $A_d^* = I_{15} + A^*(t)\Delta\mathrm{t}$ is the discretized state matrix and $B_d^* = B^*(t)\Delta\mathrm{t}$ is discretized input matrix. First, the a priori state and covariance matrix of the model were determined using equation (53), and the Kalman gain was calculated using equation (54):

$$\begin{aligned} \hat{x}^*(k|k-1) &= A_d^*\hat{x}^*(k-1) + B_d^*u^*(k-1) \\ P(k|k-1) &= A_d^*P^*(k|k-1)A_d^{*T} + Q^* \end{aligned} \tag{53}$$

$$K(k) = P^*(k|k-1)C^{*T}(C^*P^*(k|k-1)C^{*T} + R^*)^{-1} \tag{54}$$

where $Q^*$ and $R^*$ denote the noise covariance matrices. The estimated state vector $\hat{x}^*$ and covariance matrix $P^*$ at the current step are obtained using the calculated Kalman gain as described in equation (55) and equation (56), respectively.

$$\hat{x}^*(k|k) = \hat{x}^*(k|k-1) + K(k)\big(y^*(k) - C^*\hat{x}^*(k|k-1)\big) \tag{55}$$

$$P^*(k|k) = P(k|k-1) - K(k)C^*P(k|k-1) \tag{56}$$

### *B. Simulation Scheme*

The entire vehicle model simulation was conducted by assuming a vehicle moving at a certain speed on a road, where both the EKF and LSTM models were implemented in MATLAB/Simulink (solver of ode4 Runge-Kutta, sampling time of 0.01s). For the LSTM model, the eight hyperparameters mentioned in section II-A were used in this study. As previously discussed, Bayesian optimization was achieved by using prior information about the optimization process, allowing it to effectively search for new sets of hyperparameters within a predefined searching space range. Fig. 7 illustrates the progress of the optimization process over successive iterations.

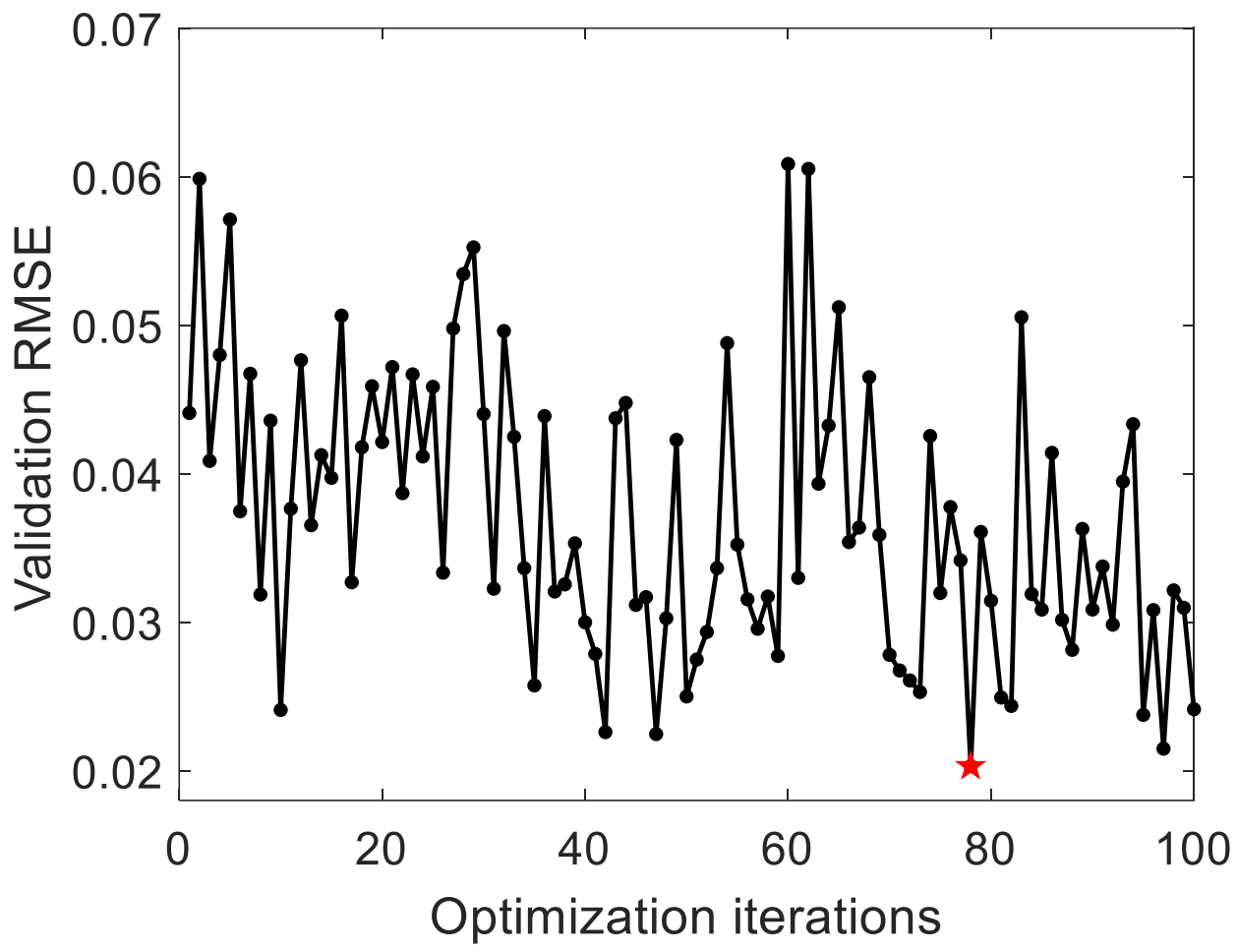

Fig. 7. Bayesian optimization process to find optimal hyperparameters. The red star-shaped point indicates the lowest validation RMSE obtained during the optimization process.

Through 100 optimization iterations, we selected the optimal hyperparameter set corresponding to the 78th interaction point with the result of minimum RMSE 0.0203.

TABLE II
SELECTED HYPER-PARAMETERS FOR BAYESIAN OPTIMIZATION FOR LSTM MODEL

| **Hyper-parameter** | **Range** | **Optimal value** |
|---|---|---|
| Maximum epochs | 500-700 | 508 |
| Validation frequency | 3-10 | 10 |
| Gradient threshold | 0.5-1.5 | 0.886 |
| Initial learning rate | 0.001-0.01 | 0.0039 |
| Learning rate drop period | 100-200 | 162 |
| Learning rate drop factor | 0.2-0.4 | 0.386 |
| Mini-batch size | 32-128 | 116 |
| Sequence length | 5000-10000 | 6553 |

For the extended Kalman filter model, initial values for $x_0$ and $P_0$ should be established before the algorithm is executed. Due to the difficulty of setting the initial values of the state vector and covariance matrix to practical values, arbitrary values were initially assigned. The noise covariance matrices value of $Q^*$ and $R^*$ were refined through a trial-and-error approach. In this study, the covariance vales of $Q^*$ and $R^*$ were selected as :

$$Q^* = diag[1000 \quad 1000 \quad 100 \quad 1 \quad 1 \quad 1 \quad 1 \quad Q_{f[1\times8]}] \times 10^{-3}$$
$$R^* = diag[10 \quad 1 \quad 10 \quad 1 \quad 1 \quad 1 \quad 1 \quad 1 \quad 1]$$

where $Q_{f[1\times8]}$ are the covariance vectors of longitudinal and lateral tire forces. Those values are selected as followed:

$$Q_{f[1\times8]} = [1000 \quad 1000 \quad 1000 \quad 1000 \quad 1 \quad 1 \quad 1 \quad 1]$$

## V. SIMULATION RESULT

### A. Estimation Performance

To evaluate the basic estimation performance of the LSTM model combined with the MPC controller, a simulation of the D-class vehicle model without noise input data was performed, assuming that the vehicle traveled at a velocity of 80 km/h for 10 s (distance of 222m). In addition, the root mean square error (RMSE), expressed as follows, was used to quantify the estimation accuracy [38].

$$RMSE(k) = \sqrt{\frac{1}{k}\sum_{i=1}^{k}\left(p(i) - \hat{p}(i)\right)^2} \qquad (57)$$

where $k$ is a time index at $t = k\Delta t$, $p$ is the true value obtained from the CarSim simulator and $\hat{p}$ is the estimated value to be compared with the true value.

The estimation results of the EKF and LSTM using the integrated MPC and DYC for four-wheel independent steering and driving (4WISD) vehicle were shown in Fig. 8 and Fig. 9, respectively. The reference used to calculate the RMSE of the longitudinal and lateral tire forces was based on the true values obtained from CarSim. As shown in Table III, the results indicate that the RMSE values of the LSTM were lower than those of the EKF in all cases, except for RF and RR in the longitudinal tire force. The substantial difference in the RMSE values between the LSTM and EKF for the estimation of lateral tire forces, which play a dominant role in lateral stability, indicates that the estimation performance of the LSTM is highly accurate.

TABLE III
COMPARISON OF RMSE FOR THE TIRE FORCES OF EACH WHEEL IN 4WISD-DYC WITH EKF 4WISD-DYC WITH LSTM

| | Longitudinal tire force RMSE (N) | | | | Lateral tire force RMSE (N) | | | |
|---|---|---|---|---|---|---|---|---|
| | FL | FR | RL | RR | FL | FR | RL | RR |
| EKF | 198.7 | **205.7** | 182.5 | **184.1** | 997.8 | 717.7 | 728.8 | 519.3 |
| LSTM | **128.3** | 210.8 | **158.9** | 196.9 | **295.1** | **287.5** | **250.5** | **137.7** |

The vehicle stability was evaluated using the $\beta$-$\dot{\beta}$ phase plane method, which relied on the side slip angle $\beta$ and its rate of change. The steering angles of each wheel of 4WISD-DYC with EKF and the area occupied in the $\beta$-$\dot{\beta}$ phase plane are significantly larger than those of 4WISD-DYC with LSTM, as shown in Fig. 10(a) and Fig. 10(b). This indicates that the 4WISD-DYC with LSTM allows more stable driving with a sufficiently lower steering angle workload than the 4WISD-DYC with EKF.

Accurate tire force estimation is required for the SMC to follow the desired yaw moment and overcome the cornering resistance that occurs during rotational vehicle motion. The torques generated by the 4WISD-DYC with the EKF and that of 4WISD-DYC with LSTM are shown in Fig. 11(a) and Fig. 11(b), respectively. The 4WISD-DYC with the EKF insufficiently satisfies the required torque for stable turning on slippery road conditions because its tire force prediction performance is lower than that of the 4WISD-DYC with LSTM.

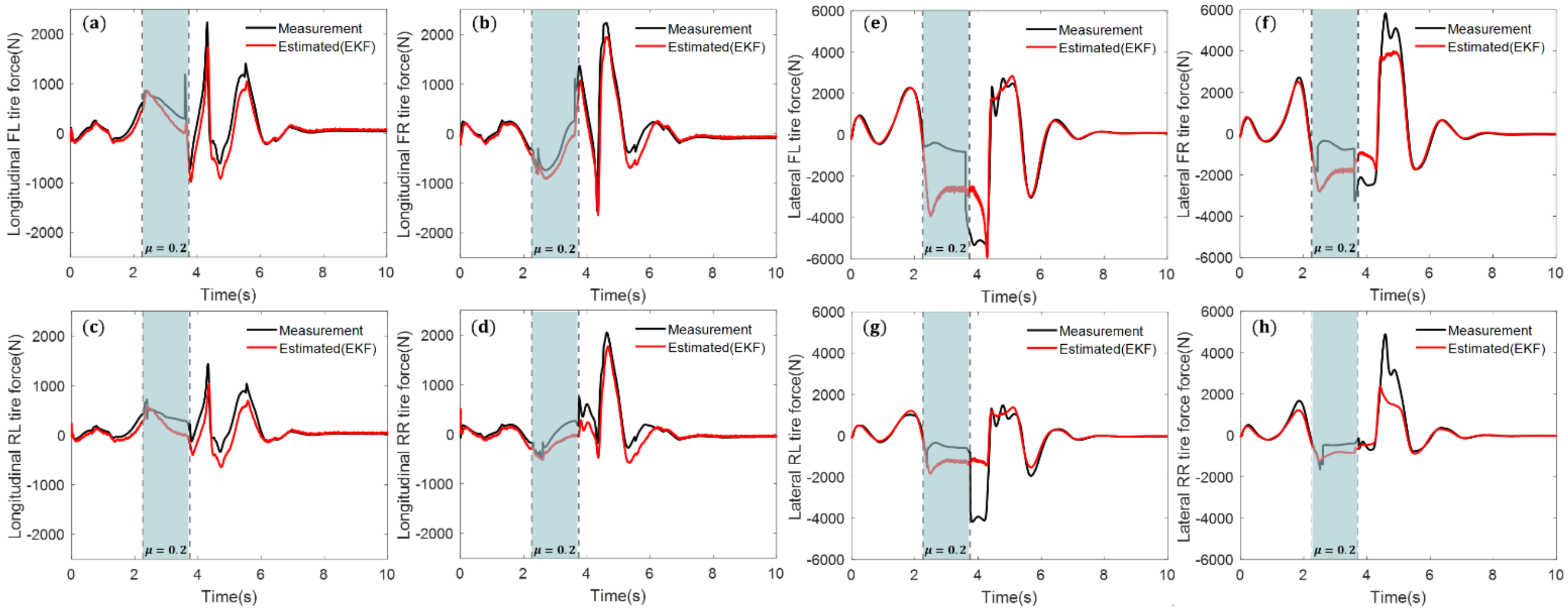

Fig. 8. Extended Kalman filter estimation result of longitudinal and lateral tire forces for each wheel; (a) longitudinal FL tire force, (b) longitudinal FR tire force, (c) longitudinal RL tire force, (d) longitudinal RR tire force, (e) lateral FL tire force, (f) lateral FR tire force, (g) lateral RL tire force, (h) lateral RR tire force.

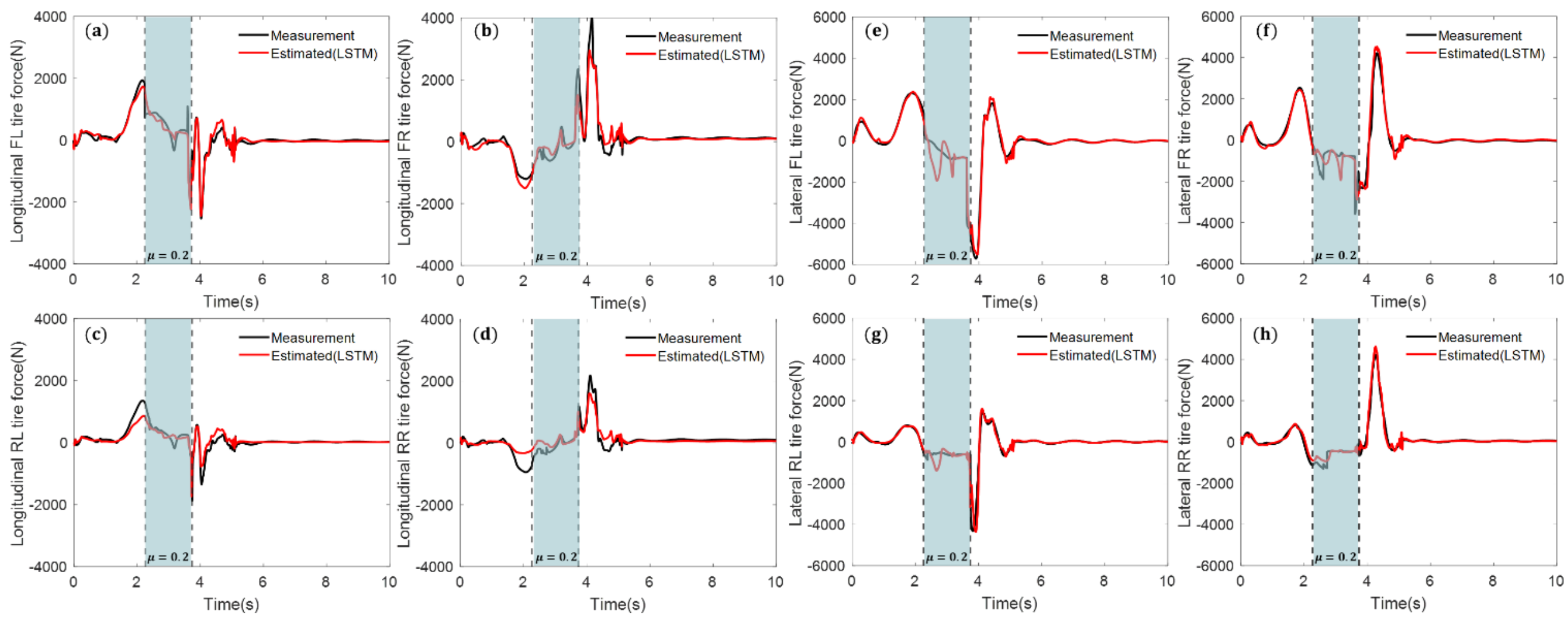

Fig. 9. LSTM estimation result of longitudinal and lateral tire forces for each wheel; (a) longitudinal FL tire force, (b) longitudinal FR tire force, (c) longitudinal RL tire force, (d) longitudinal RR tire force, (e) lateral FL tire force, (f) lateral FR tire force, (g) lateral RL tire force, (h) lateral RR tire force.

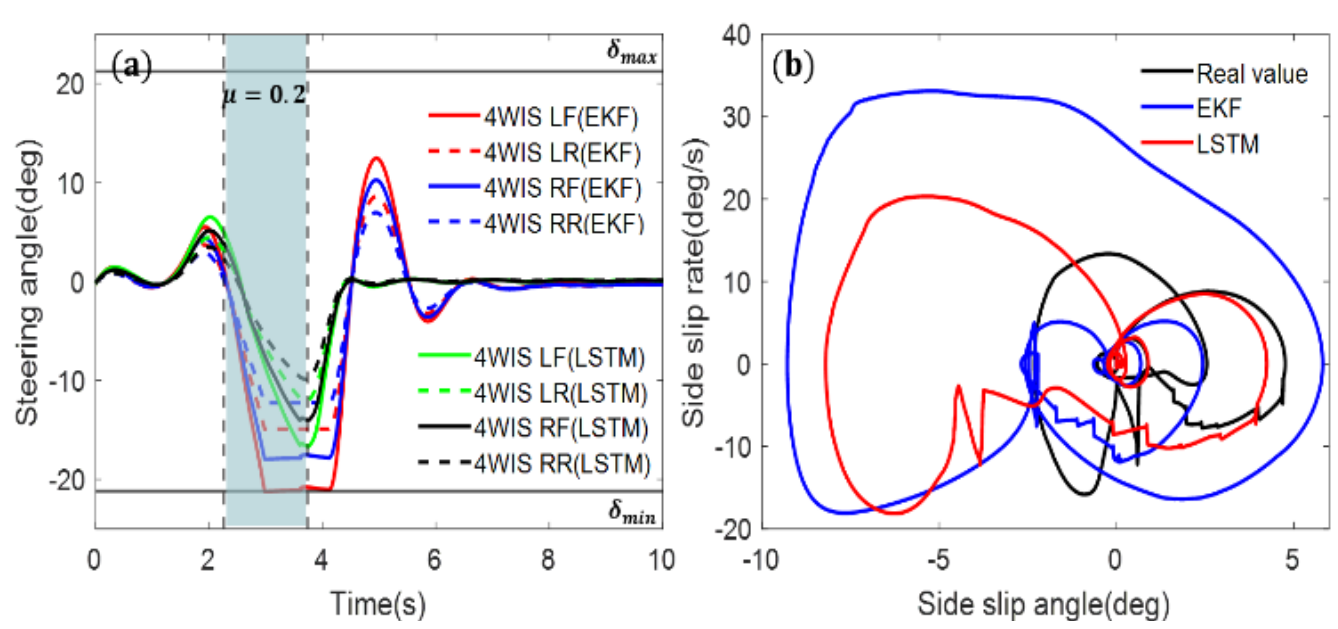

Fig. 10. (a) The steering angles of each wheel in 4WISD-DYC with EKF and 4WISD-DYC with LSTM, (b) The side slip angle and side slip rate phase plane.

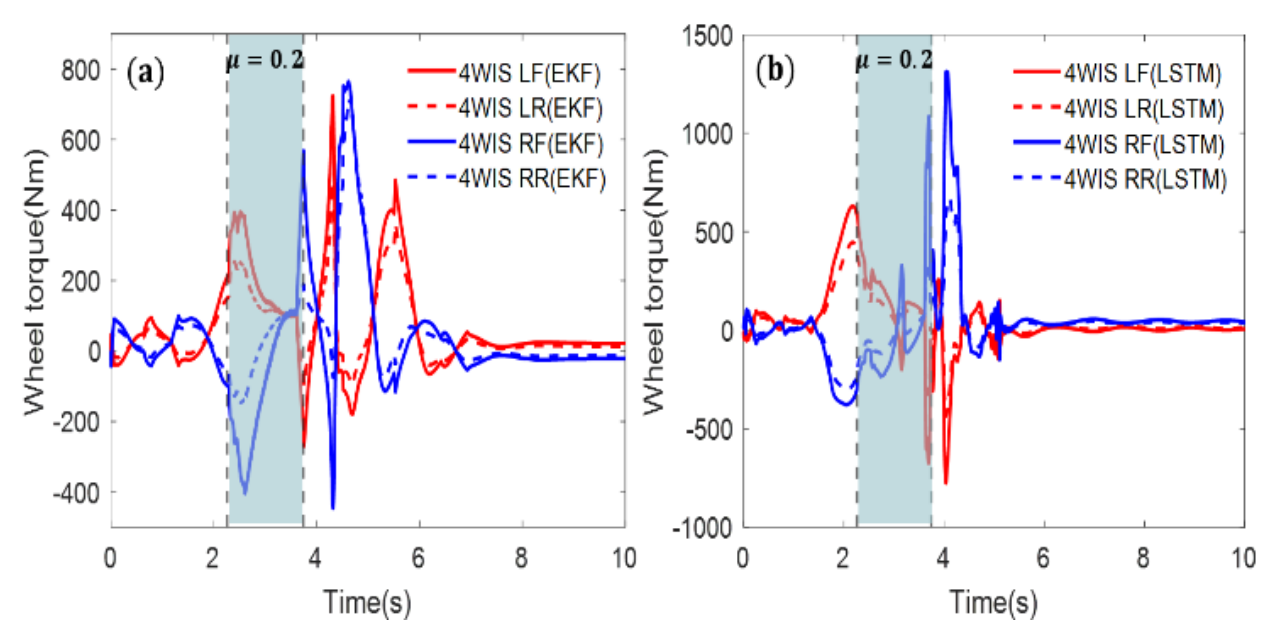

Fig. 11. Comparison of the torque of each wheel (a) 4WISD-DYC with EKF, (b) 4WISD-DYC with LSTM.

Moreover, the path trajectories of 4WISD-DYC with the EKF and 4WISD-DYC with LSTM were also compared by using the previously mentioned results, as shown in Fig. 12. To evaluate vehicle instability during the path tracking control, the maximum lateral distance to the reference path was used as the evaluation metric. The maximum lateral distance of the reference path to the road center (i.e., $Y = 0$) is 1.8 m, and for 4WISD-DYC with EKF and 4WISD-DYC with LSTM, the maximum path departure errors to the reference path are 1.28 m and 0.32 m, respectively. Therefore, the lateral departure error rates to the reference path of 4WISD-DYC with EKF and 4WISD-DYC with LSTM were 71% and 18%, respectively. This demonstrates that 4WISD-DYC with LSTM based estimator was able to robustly estimate the longitudinal and lateral tire forces to significantly improve the path tracking performance in the presence of slippery road conditions.

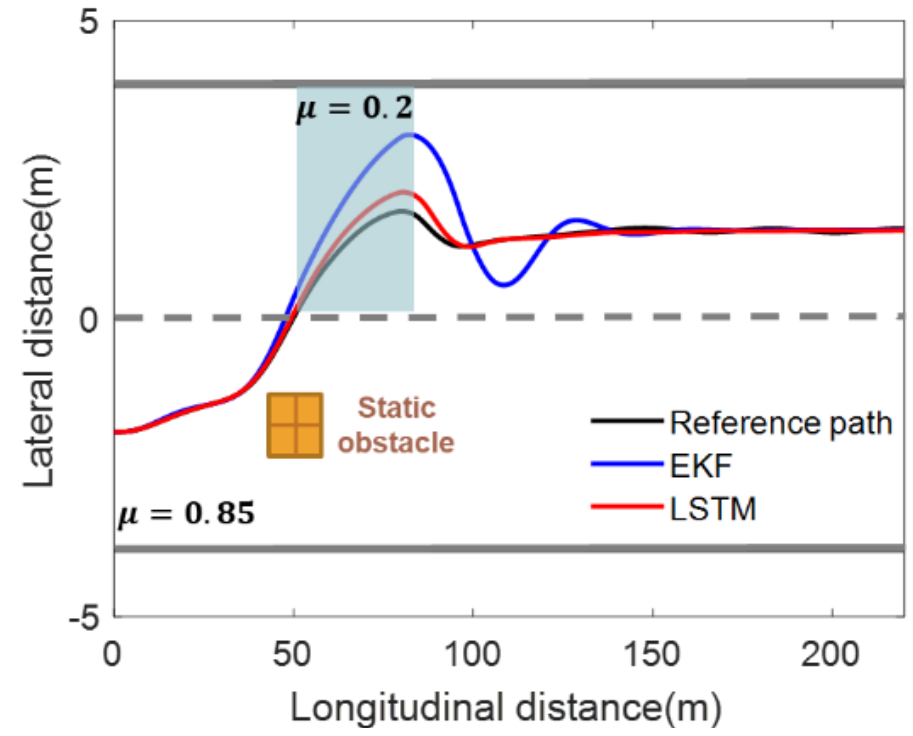

Fig. 12. Comparison of the path trajectory 4WISD-DYC with EKF and 4WISD-DYC with LSTM. (μ: 0.85 → 0.2 → 0.85)

## *B. Sensitivity Analysis*

The sensitivity of the proposed LSTM estimator combined with MPC with DYC algorithm can be analyzed through the diverse dynamic behavior due to sensor noise which are importantly considered for real world experiments. In the real world, since the data obtained by using the sensor contains electrical random noise, we thus added the measured data with white Gaussian noise. Simulations were performed at two different noise levels. First, case 1 refers to a set of contaminated noise, and second, case 2 refers to a set of more contaminated noise. For example, white Gaussian noise data with variances of 0.02 and 0.04 was added to yaw rate data, which is one of the measurement data commonly used in LSTM and EKF estimators, as shown in Fig. 13. The variance values of the other measured input data required for the LSTM and EKF tire force estimators were also added as shown in Table VI and Table V.

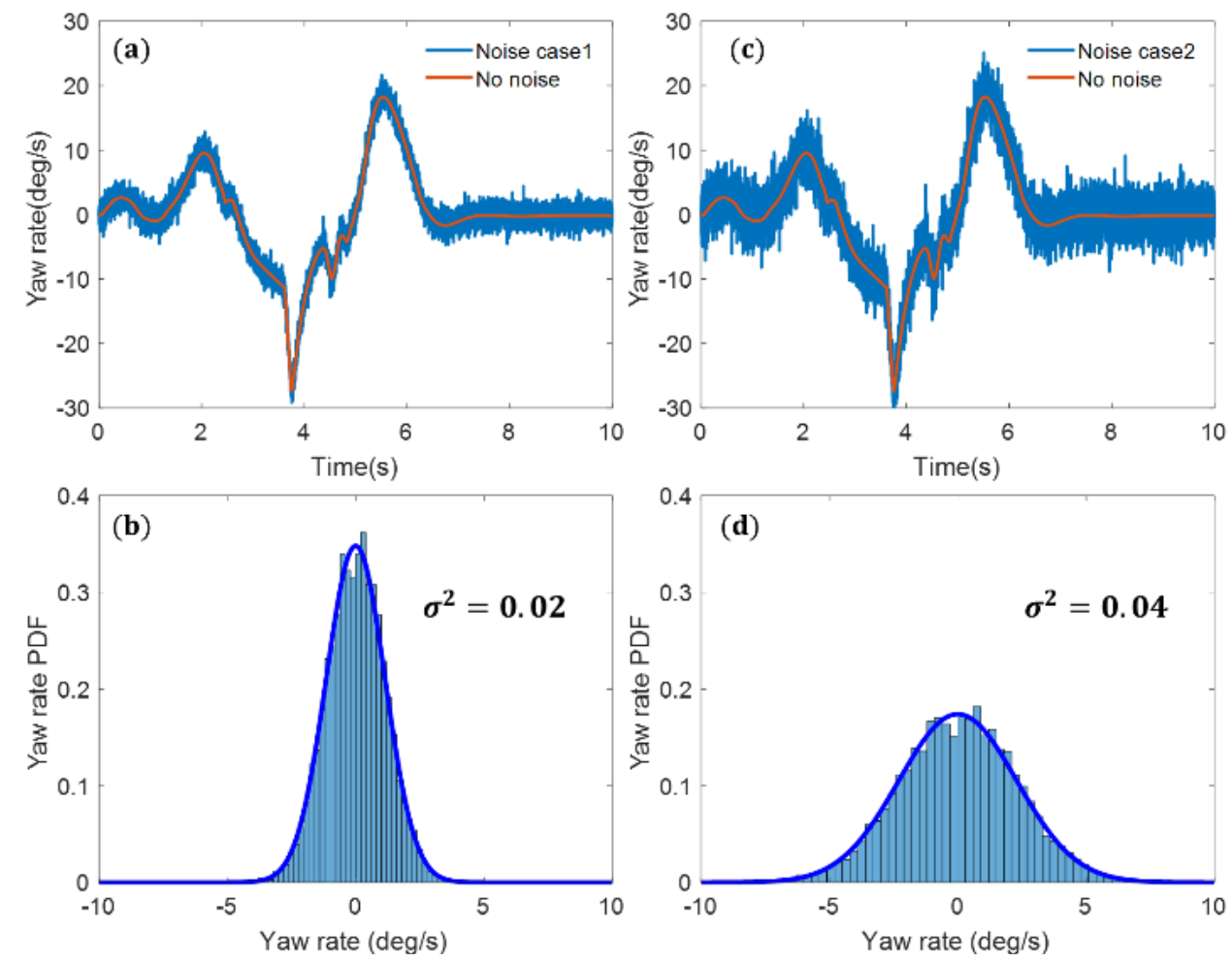

Fig. 13. White Gaussian random noise added yaw rate sensor data; (a)-(b) contaminated, (c)-(d) two times largely contaminated than (a)-(b).

TABLE VI
THE VARIANCE OF MEASURED SENSOR NOISE VALUES USED FOR LSTM

| Measurement sensor data | $\dot{\zeta}(deg/s)$ | | $\dot{\theta}(deg/s)$ | | $\gamma(deg/s)$ | | $\zeta(deg)$ | | $\varphi(deg)$ | |
|---|---|---|---|---|---|---|---|---|---|---|
| Variance of noise($\sigma^2$) | case1 | case2 | case1 | case2 | case1 | case2 | case1 | case2 | case1 | case2 |
| | 0.5 | 1 | 0.031 | 0.062 | 0.25 | 0.5 | 0.125 | 0.25 | 0.125 | 0.25 |

TABLE V
THE VARIANCE OF MEASURED SENSOR NOISE VALUES USED FOR EKF

| Measurement sensor data | $v_x(m/s)$ | | $v_y(m/s)$ | | $a_x(g)$ | | $a_y(g)$ | | $\gamma(rad/s)$ | | $w_{FL}(rpm)$ | | $w_{FR}(rpm)$ | | $w_{RL}(rpm)$ | | $w_{RR}(rpm)$ | |
|---|---|---|---|---|---|---|---|---|---|---|---|---|---|---|---|---|---|---|
| Variance of noise($\sigma^2$) | case1 | Case2 | case1 | case2 | case1 | case2 | case1 | case2 | case1 | case2 | case1 | case2 | case1 | case2 | case1 | case2 | case1 | case2 |
| | 0.02 | 0.04 | 0.2 | 0.4 | 0.02 | 0.04 | 0.02 | 0.04 | 0.02 | 0.04 | 6 | 12 | 6 | 12 | 6 | 12 | 6 | 12 |

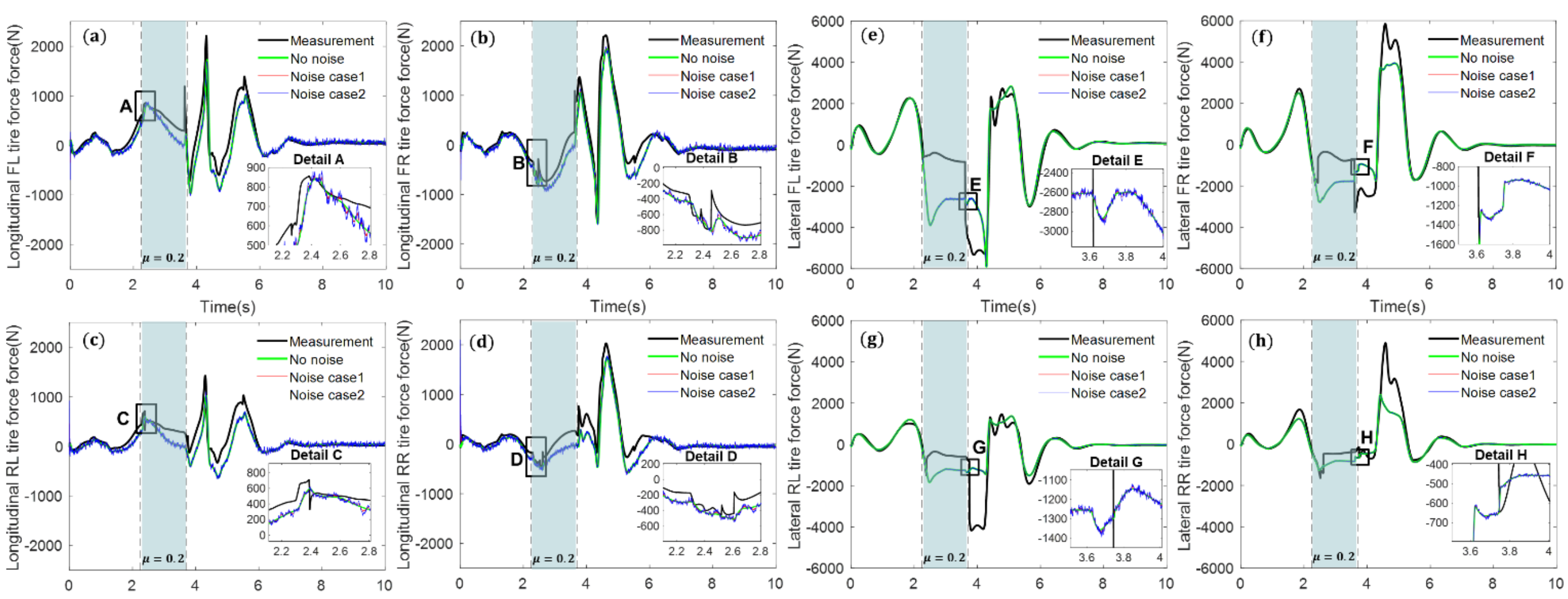

Fig. 14. Extended Kalman filter estimation result of longitudinal and lateral tire forces for each wheel according to different sensor noise levels; (a) longitudinal FL tire force, (b) longitudinal FR tire force, (c) longitudinal RL tire force, (d) longitudinal RR tire force, (e) lateral FL tire force, (f) lateral FR tire force, (g) lateral RL tire force, (h) lateral RR tire force.

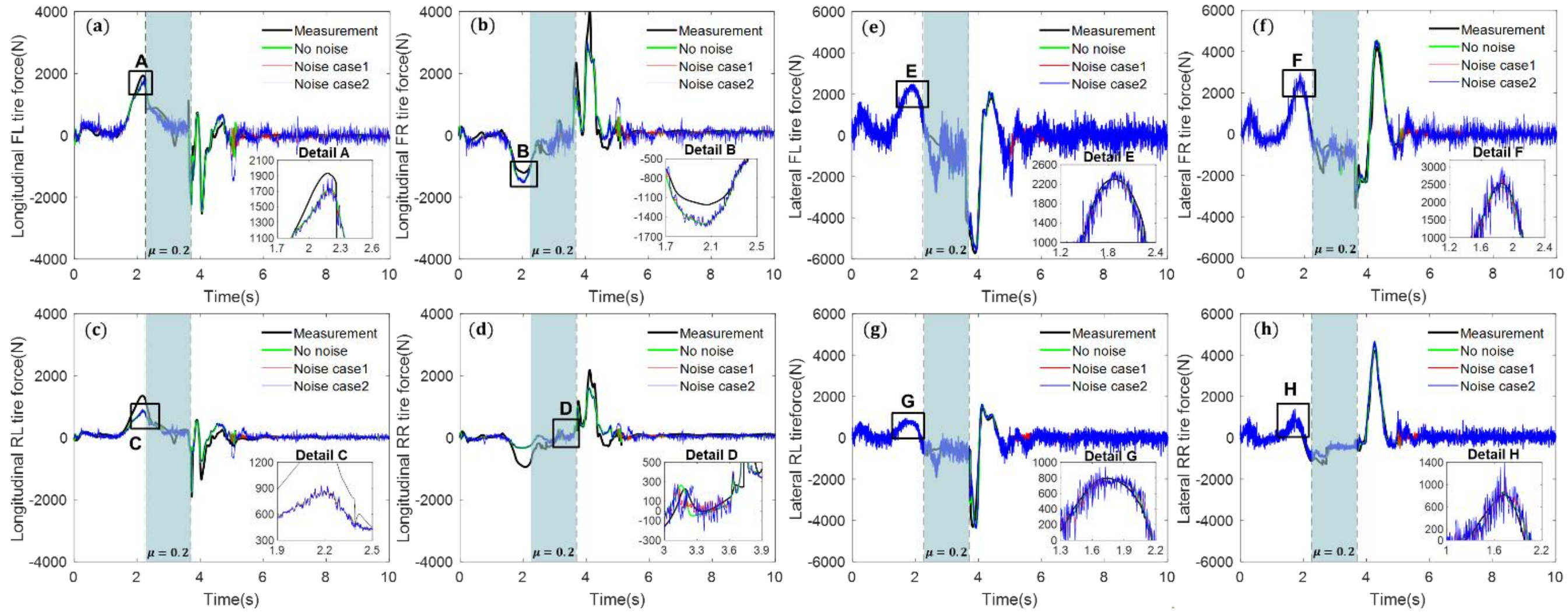

Fig. 15. LSTM estimation result of longitudinal and lateral tire forces for each wheel according to different sensor noise levels; (a) longitudinal FL tire force, (b) longitudinal FR tire force, (c) longitudinal RL tire force, (d) longitudinal RR tire force, (e) lateral FL tire force, (f) lateral FR tire force, (g) lateral RL tire force, (h) lateral RR tire force.

The simulation results of longitudinal and lateral tire forces, estimated by sensor data with different noise levels using EKF and LSTM tire force estimators, are shown in Fig. 14 and Fig. 15, respectively. To further investigate the sensitivity of the proposed LSTM model under different sensor noise levels, the relative error for nominal (i.e., normalized performance scale) was quantitatively calculated and compared with the EKF tire force estimator, as shown in equation (58).

$$\text{Relative Error} = \frac{|RMSE_p - RMSE_n|}{RMSE_n} \tag{58}$$

where $RMSE_p$ is the RMSE value perturbed by noise, and $RMSE_n$ is the RMSE of the no noise data.

Fig. 16 shows the relative error of the longitudinal and lateral tire force for all four wheels under the use of measurement data including noise. In all noise cases, it can be seen that LSTM is more sensitive to noise in estimating tire force than EKF. The Kalman filter can track the covariance of noise, allowing it to assess the level of uncertainty and autonomously compensate for a certain level of noise. On the other hand, LSTM tends to respond to new patterns or structures that have not been seen in the course of learning. Therefore, given new data containing noise, the LSTM is sensitive to noise because it recognizes that the noise is also likely to be some important information and makes predictions.

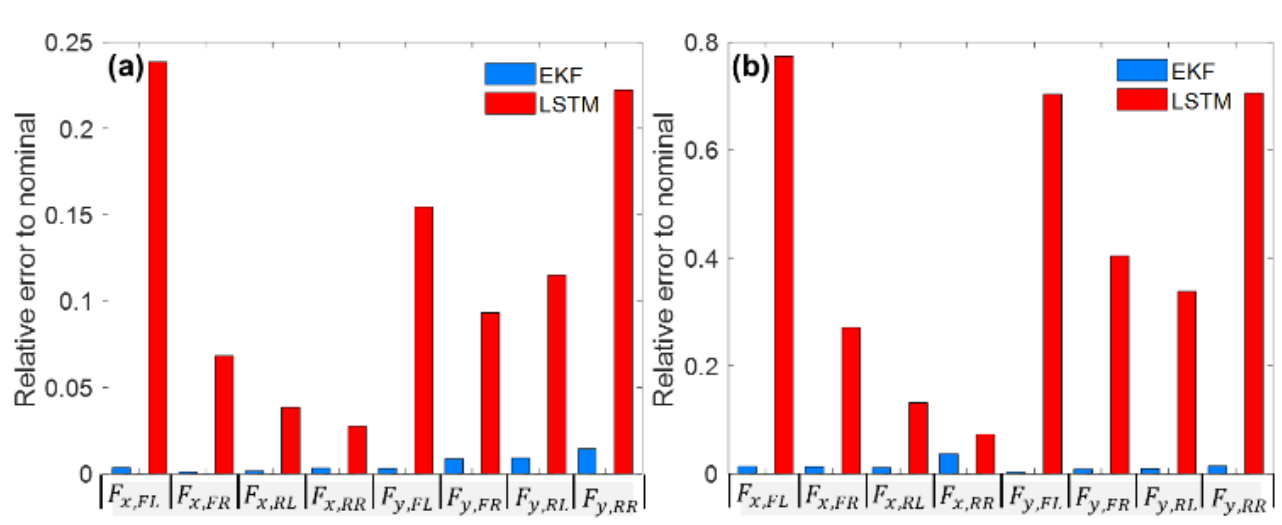

Fig. 16. Relative error to nominal under measured noise cases; (a) noise case1 (b) noise case2.

Fig. 17 shows the tracking results of the LSTM and EKF tire estimator-based MPC controller using measurement data including different noise level. Due to the characteristics of the previously mentioned LSTM and Kalman filter with respect to noise, it can be observed from detail A in Fig. 17(a) that a larger variance in path tracking error is resulted compared to detail B in Fig. 17(b). However, LSTM is sensitive to measurement data including noise, but has more accurate estimation performance than EKF, so LSTM is superior in path tracking performance.

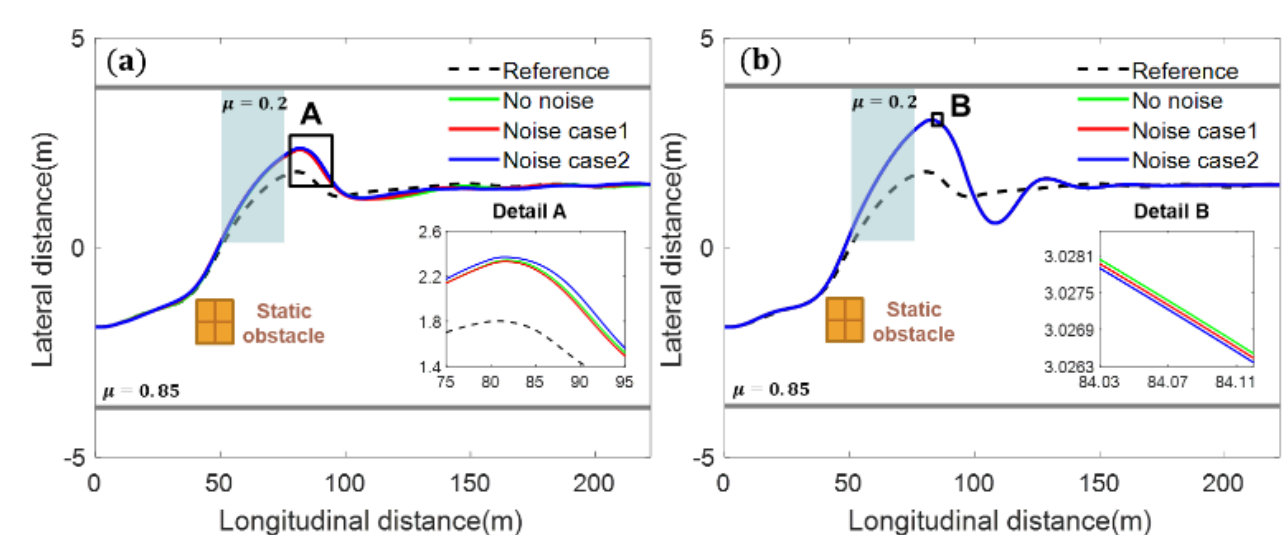

Fig. 17. Comparison of 4WISD system simulation results for different noise levels; (a) vehicle trajectories with LSTM based tire force estimator, (b) vehicle trajectories with EKF tire force estimator.

## VI. Conclusion

This study investigated that the path tracking performance could be significantly improved using the proposed LSTM-based tire force estimator combined with MPC and DYC in a four-wheel independent steering and driving vehicle under slippery road conditions compared to using the EKF tire force estimator. This study also noted that the LSTM based path tracking controller could simultaneously estimate the longitudinal and lateral tire forces using only the IMU sensor for measuring five types of sprung mass sensor data, whereas an extended Kalman filter requires nine measurement data. Therefore, the proposed approach can be highly promising for autonomous driving systems in terms of both cost and reliability. This study also describes that the EKF performs a relatively good estimation performance in terms of the RMSE of the predicted tire force values. However, the vehicle model in the EKF is assumed to be a linear vehicle and tire model through linearization, which results in a degraded estimation performance compared with the LSTM, which fully considers the nonlinearity of the vehicle tires, suspension, and complex mechanical joints. Therefore, by calculating the accurate tire forces through LSTM and estimating vehicle lateral velocity from it, not only was the accuracy of the overall vehicle system model improved, but also the performance of the controller was enhanced. Although the LSTM-based tire force estimator is sensitive to noise, it has more accurate estimation performance than EKF, ultimately enabling safe driving through accurate path tracking. Thus, this study convinces that the LSTM based tire force estimator learned through Bayesian optimization, which has an accurate estimation performance even under rapidly changing environmental scenarios, can be a deterministic approach offering more advantages for improving path tracking and stability performance in the presence of nonlinear vehicle parameters and diverse driving environments.